\documentclass{article}

\usepackage{microtype}
\usepackage{graphicx}
\usepackage{subfigure}
\usepackage{booktabs} 
\usepackage{balance}

\usepackage{hyperref}

\usepackage[accepted]{icml2024}

\usepackage{amsmath}
\usepackage{amssymb}
\usepackage{mathtools}
\usepackage{amsthm}
\usepackage{bm}

\usepackage{times}
\usepackage{epsfig}
\usepackage{multirow}
\usepackage{bbold}
\usepackage{threeparttable}
\usepackage{array}

\usepackage{bbding}
\usepackage{makecell}
\usepackage{balance}  
\usepackage{diagbox}
\usepackage{colortbl}
\usepackage{pifont}

\newcommand{\figref}[1]{Figure~\ref{fig:#1}}
\newcommand{\tabref}[1]{Table~\ref{tab:#1}}
\newcommand{\eqnref}[1]{Eq.~\eqref{eq:#1}}

\newcommand{\Paragraph}[1]{\noindent\textbf{#1}}

\newlength\secmargin
\newlength\subsecmargin
\newlength\paramargin
\newlength\figmargin
\newlength\eqmargin
\definecolor{Gray}{rgb}{0.5, 0.5, 0.5}  

\usepackage[capitalize,noabbrev]{cleveref}

\theoremstyle{plain}

\theoremstyle{definition}

\theoremstyle{remark}

\usepackage[textsize=tiny]{todonotes}

\icmltitlerunning{Fast-Slow Test-Time Adaptation for Online Vision-and-Language Navigation}

\begin{document}

\twocolumn[
\icmltitle{Fast-Slow Test-Time Adaptation for \\
           Online Vision-and-Language Navigation}



\icmlsetsymbol{equal}{*}

\begin{icmlauthorlist}
	\icmlauthor{Junyu Gao}{casia,ucas}
	\icmlauthor{Xuan Yao}{casia,ucas}
	\icmlauthor{Changsheng Xu}{casia,ucas,pcl}
\end{icmlauthorlist}

\icmlaffiliation{casia}{State Key Laboratory of Multimodal Artificial Intelligence Systems (MAIS), Institute of Automation, Chinese Academy of Sciences (CASIA)}
\icmlaffiliation{ucas}{School of Artificial Intelligence, University of Chinese Academy of Sciences (UCAS)}
\icmlaffiliation{pcl}{Peng Cheng Laboratory, ShenZhen, China}


\icmlkeywords{Machine Learning, ICML}

\vskip 0.3in
]



\printAffiliationsAndNotice{}

\begin{abstract}

The ability to accurately comprehend natural language instructions and navigate to the target location is essential for an embodied agent. Such agents are typically required to execute user instructions in an online manner, leading us to explore the use of unlabeled test samples for effective online model adaptation. However, for online Vision-and-Language Navigation (VLN), due to the intrinsic nature of inter-sample online instruction execution and intra-sample multi-step action decision, frequent updates can result in drastic changes in model parameters, while occasional updates can make the model ill-equipped to handle dynamically changing environments. Therefore, we propose a Fast-Slow Test-Time Adaptation (FSTTA) approach for online VLN by performing joint decomposition-accumulation analysis for both gradients and parameters in a unified framework. 
Extensive experiments show that our method obtains impressive performance gains on four popular benchmarks. Code is available at \url{https://github.com/Feliciaxyao/ICML2024-FSTTA}.

\end{abstract}

\section{Introduction}
\label{sec:intro}
Developing intelligent agents capable of 
adhering to 
human directives remains a significant 
challenge in embodied AI. Recently, Vision-and-Language Navigation (VLN)~\cite{anderson2018vision,qi2020reverie,chen2022think,yang2023behavioral}, which requires an agent to comprehend natural language instructions and subsequently perform 
proper actions to navigate to the target location, serves as a useful 
platform for examining the instruction-following ability.

\begin{figure}[t!]
	\centering
	\includegraphics[width=1\linewidth]{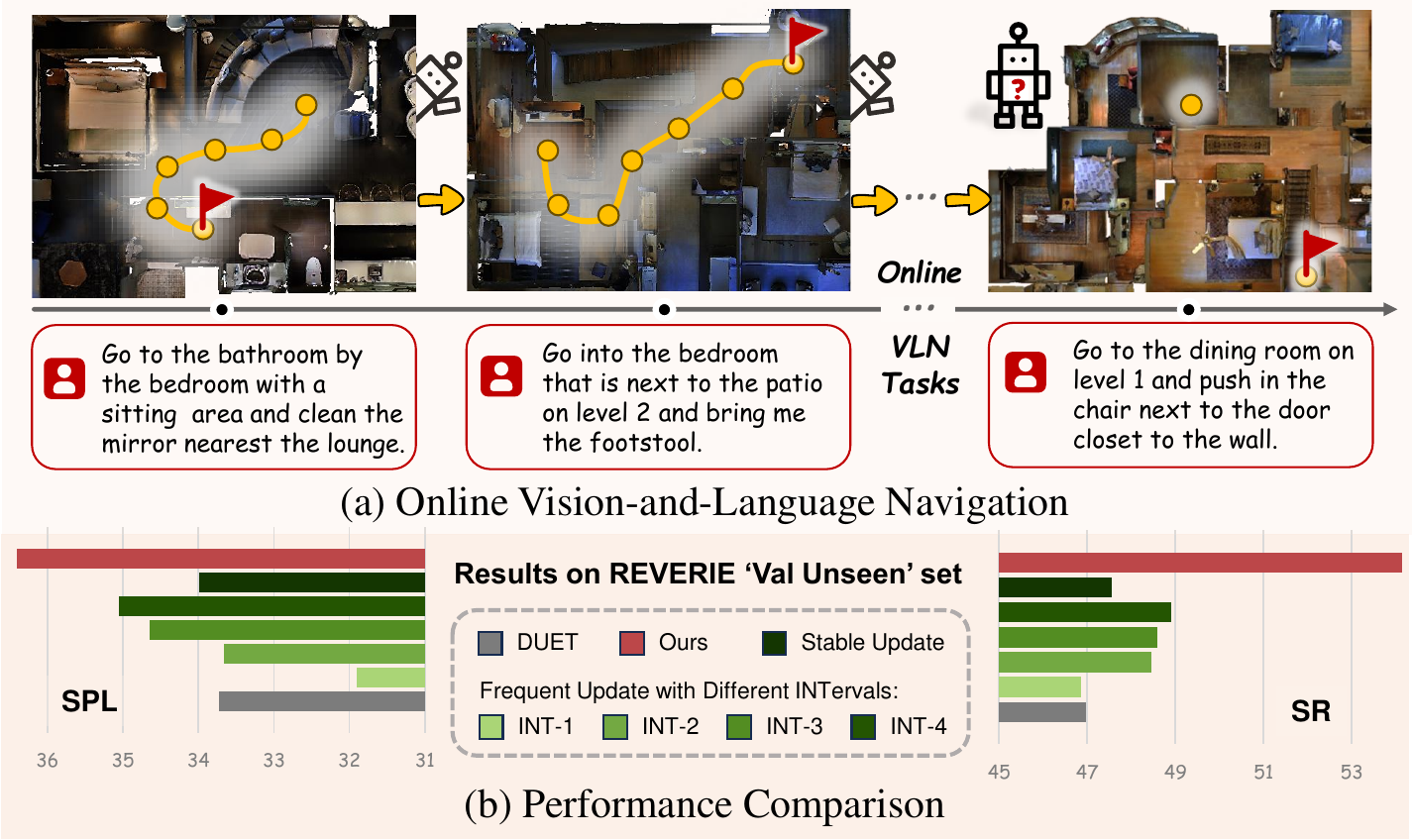}
	\caption{
		(a) Illustration of online VLN. (b) Comparison between TTA strategies on REVERIE~\cite{qi2020reverie} validation unseen set using SPL and SR metrics. 
		`DUET'~\cite{chen2022think} is the base model, `Frequent Update' means updating at certain intervals within each sample, `Stable Update' refers to initializing with the original base model for each sample and using its best intra-sample update interval INT=1. All these strategies adopt TENT~\cite{wang2020tent} for model updates. The results show that overly fast or overly slow TTA fail to achieve significant improvements.
	}
	\label{fig:motivation}
\end{figure}



In practical applications, a trained VLN agent is required to execute user instructions in various environments at different times in an online manner, as depicted in~\figref{motivation}(a). However, owing to disparities in environmental factors, such as distinct room types and objects, the trained agent inevitably confront data discrepancies during online testing~\cite{gu2022vision,Guhur2021AirbertIP}. 
This raises an important question: \emph{can an agent accumulate experience and enhance its capabilities while executing instructions?} 
However, due to the lack of annotation, updating the model via supervised online learning is impractical. In addition, other learning paradigms like unsupervised domain adaptation or semi-supervised learning are also infeasible, given considerations for the issues of execution efficiency and privacy protection.


Recently, online Test-Time Adaptation (TTA)~\cite{liang2023comprehensive,niu2023towards,wang2022continual} has been recognized as an effective technique of online model updating by leveraging unlabeled test samples. 
Although prevailing TTA methods can be integrated into VLN models with certain alterations, this direct application cannot well handle the \emph{adaptability-stability dilemma} of models due to the intrinsic nature of inter-sample online instruction execution and intra-sample multi-step action decision.
Specifically, unlike traditional classification tasks where a single TTA operation is sufficient for each test sample, as illustrated in~\figref{motivation}(a), online VLN requires an agent to perform a sequence of actions within a single test sample and to execute various samples (instructions) in an online manner.
On one hand, while conducting TTA at every (or a few) action steps enables rapid agent adaptation to dynamic environments, frequent model updates may introduce significant model alterations, potentially causing cumulative errors and catastrophic forgetting~\cite{wang2022continual,niu2022efficient,song2023ecotta}, thus compromising model stability during testing. On the other hand, initializing the same model for stable TTA in each test sample may hinder the model's ability to adaptively learn experience from 
historical test samples, thereby impeding its potential for achieving superior performance.~\figref{motivation}(b) shows that  both overly fast or overly slow model updates fail to achieve significant performance improvements.

To tackle the above issues, we propose a Fast-Slow Test-Time Adaptation (FSTTA) method for online VLN tasks. 
Built upon a unified gradient-parameter decomposition-accumulation framework, our approach consists of a fast update phase and a slow update phase, 
pursuing a balance between adaptability and stability in online model updating. Specifically, with a test-time training objective, such as entropy minimization~\cite{wang2020tent}, we can derive  gradients at each action step in the fast update phase.
However, due to the unsupervised nature of TTA, these gradients inevitably contain noise information.
Using these gradients for model update can interfere with the adaptability,
especially when the update is frequently invoked.
Therefore, we establish a local coordinate system to find a reliable optimization direction by periodically analyzing the gradients generated during the recent multi-step navigation process.
After a certain number of fast updates, the model parameters (also called model state)  
are recorded. 
To further mitigate the issues of cumulative errors and catastrophic forgetting that may result from excessively frequent model updates, during the slow update phase, we revert the model to its historical state and conduct a decomposition-accumulation analysis on the parameter variation trajectory for a direct model update. 
Both phases are performed alternately during testing to balance the adaptability and stability of the model. As shown in~\figref{motivation}(b), the proposed method achieves significant improvement against other model update strategies.

Our contributions can be summarized as follows: (1) Considering the characteristics of inter-sample online instruction-execution and intra-sample multi-step action-execution, 
we explore the online VLN task and propose a fast-slow test-time adaptation (FSTTA) method for effective navigation and model updating. 
(2) Based on a unified decomposition-accumulation framework for both gradients and parameters, our method ensures swift model adaptability to environmental changes in the short-term fast update phase, while preserves stability throughout the long-term slow update phase. (3) Our FSTTA elevates the performance of several leading VLN models on four popular benchmarks. 
When applied to the notable DUET model~\cite{chen2022think}, it yields a performance boost of over $5\%$ on the representative discrete/continuous benchmarks REVERIE/R2R-CE.
Furthermore, our method shows superior results compared to other premier TTA techniques.

\section{Related Work}
\label{sec:rw}
\noindent\textbf{Vision-and-Language Navigation (VLN).}
Most existing methods facilitate VLN research by developing powerful techniques for model training, including: \textbf{(i)} Designing advanced network architectures. 
Early VLN models utilize LSTMs with various attention mechanisms~\cite{anderson2018vision,fried2018speaker,Hong2020LanguageAV} while recent ones~\cite{hao2020towards,hong2021vln,chen2021history,lin2022multimodal,chen2022think,huo2023geovln} resort to the more popular transformer-based methods for multi-modal pre-training. Other architectures are also explored such as graph neural networks~\cite{zhu2021soon} and parameter-efficient adapter~\cite{qiao2023vlnpetl}. \textbf{(ii)} Adopting various training paradigms 
such as reinforcement and imitation  learning~\cite{nguyen2019vision,tan2019learning,wang2019reinforced,gao2023adaptive}. Moreover, to estimate the completeness of instruction following and decide when to conduct backtracking, progress monitoring~\cite{ma2019self,zhu2020vision} and back-tracking~\cite{ke2019tactical,ma2019regretful} 
are also employed to promote training process. \textbf{(iii)} Performing data augmentation for training a stronger model. In recent years, more and more large-scale benchmarks are established via collecting human annotations~\cite{ku2020room,zhu2021soon,ramrakhya2022habitat} or creating new environments~\cite{qi2020reverie,chen2022learning}.
Other approaches explore techniques such as mixup and synthesis~\cite{Liu2021VisionLanguageNW,kamath2023new}, style transfer~\cite{li2022envedit}, or future-view image semantics~\cite{li2023improving} for data augmentation. \textbf{(iv)} Leveraging additional information for boosting model capacity. Since the goal of VLN is to navigate in photo-realistic environments, there are many kinds of information in the world that can be used such as knowledge~\cite{li2023kerm}, 3D scene geometry~\cite{Liu2023BirdsEyeViewSG,wang2023gridmm}, and landmarks~\cite{wang2022less,cui2023grounded}. 

Although the above methods have made significant progress in training effective models, they overlook the utilization of test data during the online VLN process. In real-world applications, an agent is required to continuously execute user instructions at different times. The ability of an agent to accumulate experience during this process would greatly enhance its practical value. 
Note that~\cite{lu2022anticipating} first explored test-time adaptation for VLN. However, this approach does not perform model update in an online manner and overlook the balance between adaptability-stability.


\noindent\textbf{Online Test-time Adaptation (TTA).}  TTA  allows models to adapt the test data in an online and unsupervised manner~\cite{liang2023comprehensive,lim2022ttn,lee2023towards}. 
Existing TTA methods generally rely on batch normalization calibration~\cite{mirza2022norm,zhao2023delta,gong2022note}, entropy minimization~\cite{wang2020tent,niu2023towards,tang2023neuro}, auxiliary self-supervised task or data regularization~\cite{sun2020test,boudiaf2022parameter,zhang2022memo} to acquire useful information for reducing the domain gap between training and testing data. 
To stabilize adaptation in continuously changing data distribution, recently, continual test-time adaptation~\cite{wang2022continual,niu2022efficient,song2023ecotta,dobler2023robust,yuan2023robust,Liu2023ViDAHV}, as a more practical setting, 
has been tentatively explored for addressing the cumulative errors and catastrophic forgetting issues.
Until now, test-time adaptation has been preliminarily explored in some sequential data analysis fields such as action recognition~\cite{lin2023video} and video classification~\cite{yi2022temporal}. However, they overlook the joint inter- and intra-sample structure in TTA of sequential data.

\noindent\textbf{Gradient-based Methods.} 
Gradients are central to modern SGD-based deep learning algorithms. To date, gradient analysis research has predominantly focused on domain generalization (DG)~\cite{mansilla2021domain,lew2023gradient,wang2023sharpness,rame2022fishr,wang2022pgrad,tian2023trainable}, due to the negative impact of conflicting gradients from multiple domains on model optimization. Pioneering works~\cite{du2018adapting,yu2020gradient,mansilla2021domain} perform gradient surgery at the backpropagation phase via various 
strategies such as normal plane projection~\cite{yu2020gradient} and consensus learning~\cite{mansilla2021domain}. 
Other approaches resort to gradient 
agreement regularization for refining the optimization direction by leveraging sharpness~\cite{wang2023sharpness} or similarity~\cite{shi2021gradient,rame2022fishr} measurements.
Different from the above strategies that only consider a single-phase gradient surgery in DG, we jointly analyze the gradient-parameter states for a two-phase (fast-slow) TTA in the VLN task.


\section{Our Approach}
\label{sec:our}

\begin{figure*}[t!]
	\centering
	\includegraphics[width=1\linewidth]{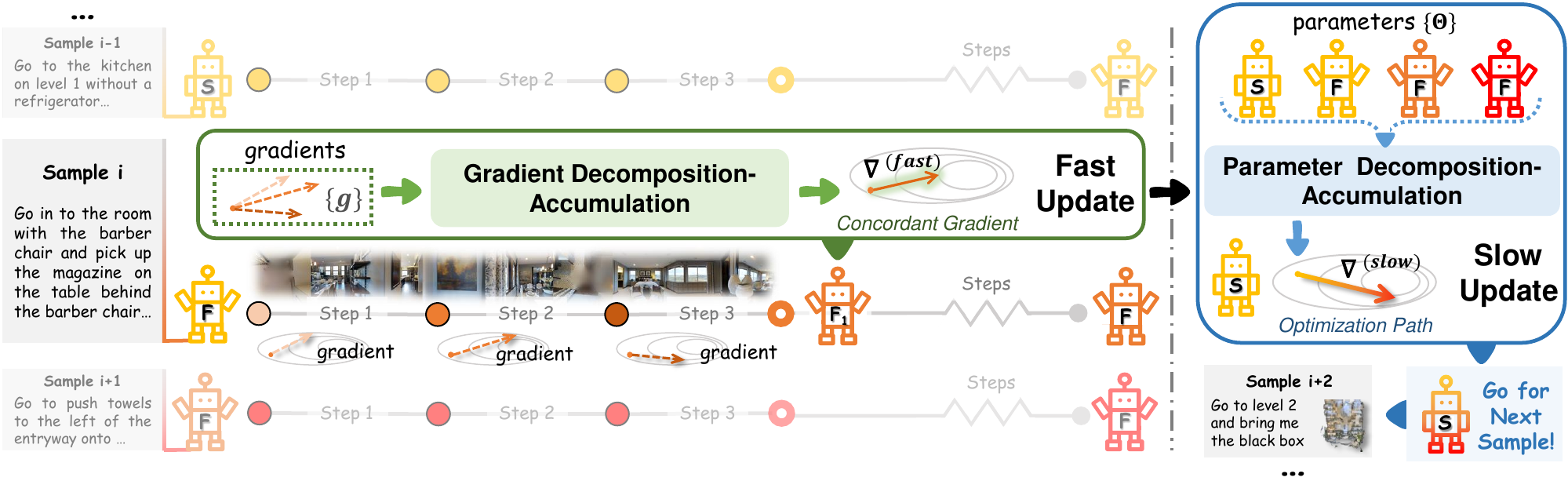}
	\caption{
		Overall framework of the proposed Fast-Slow Test-Time Adaptation (FSTTA) for online VLN. In the fast update phase, taking `Sample i' as an example, the model periodically analyzes the gradients ($\{\bm g\}$) generated during the recent multi-step navigation and performs a gradient decomposition-accumulation analysis to pinpoint a concordant direction for model update. After a certain number of fast updates, historical model parameters ($\{\bm \Theta\}$) are recorded. In the slow update phase, we revert the model to its historical state and conduct a parameter decomposition-accumulation analysis to learn an optimization path for direct parameter modulation. Note that `F', `S' in the robots means the model parameters after fast and slow updates. `$\text{F}_1$' indicates the first fast update within a test sample. 
	}
	\label{fig:framework}
\end{figure*}

\vspace{\subsecmargin}

\Paragraph{Problem Setup and VLN Base Model.}
Given a natural language instruction $\bm I$,
the VLN task requires an agent to find the target viewpoint through the environment by executing a series of actions. 
During the navigation process, an undirected exploration graph $\mathcal{G}_t = (\mathcal{V}_t, \mathcal{E}_t)$ is progressively constructed, where $\mathcal{V}_t$ denotes navigable nodes, $\mathcal{E}_t$ indicates the connectivity edges, $t$ is the current timestep. 
Note that a `STOP' node is added to this graph to indicate a stop action, and we connect it with all other nodes. 
At this moment, 
the agent receives a panoramic view that contains 36 single images. 
The panorama is represented by the image features $\mathcal{R}_t$ and their object features $\mathcal{O}_t$, where these features can be extracted by pre-trained vision transformers (ViT)~\cite{dosovitskiy2020image,chen2022think,li2023kerm}. 
To accomplish the instruction, the agent needs to predict probabilities for the currently navigable nodes and select the most possible one as the next movement action. The probabilities can be predicted as:
\begin{equation}\label{eq:action_score}
	\bm s_{t} =  \phi\left(\bm I, \mathcal{R}_t, \mathcal{O}_t, \mathcal{H}_t; \bm{\Theta}\right),~~ \bm s_{t} \in \mathbb{R}^{|\mathcal{V}_t|}
\end{equation}
where $\mathcal{H}_t$ indicates the history information that encodes the observed visual features and performed actions~\cite{chen2022think,qiao2023vlnpetl}. $\phi(\cdot)$ is the VLN base model such as the dual-scale graph transformer~\cite{chen2022think,chen2022learning}, $\bm{\Theta}$ is the learnable model parameters. 


\Paragraph{Framework Overview.} In this paper, we devote to 
adjusting the VLN base model during testing process within an online and unsupervised manner. Our FSTTA framework is illustrated in ~\figref{framework}. For each sample, at timestep $t$, we employ the commonly adopted entropy minimization objective~\cite{wang2020tent,niu2023towards} for test-time adaption, which aims to reduce the entropy of the probabilities over the current navigable nodes:

\begin{equation}\label{eq:loss}
	\mathcal{L} ({\bm s}_{t}; \bm{\Theta}) = -\sum\nolimits_{i} {\bm s}_{t,i}\log({\bm s}_{t,i}).
\end{equation}

During the optimization of the above objective, gradients are back-propagated for updating the model's parameters. However, updating the whole base model is computationally infeasible. As a result, we only consider a small portion of the model parameters for gradient calculation. 
Since affine parameters in normalization layers capturing data distribution information, numerous TTA methods opt to update these parameters for adaption~\cite{wang2020tent,niu2023towards,liang2023comprehensive}. In this paper, we employ the model's a few final layer-norm operations for TTA and maintain other parameters frozen. For brevity, we still use the symbol $\bm \Theta$ to represent these parameters to be updated in the following sections, $\bm \Theta \in \mathbb{R}^{D}$. 
Targeting at fully leveraging the gradient and parameter information, under a unified decomposition-accumulation analysis framework, we propose an effective two-phase adaptation for fast and slow model updates.

\vspace{\subsecmargin}
\subsection{Fast Update via Gradient Analysis}\label{sec:fup}
\vspace{\subsecmargin}

At timestep $t$ in one navigation process, 
the agent is required to select an action (navigable node) by using the predicted score ${\bm s}_{t}$. With ${\bm s}_{t}$, we can calculate the TTA loss (\eqnref{loss}) and then derive the gradient of the model parameters $\bm{\Theta}$ as: $\bm g_t = \nabla \mathcal{L} ({\bm s}_{t};\bm{\Theta})$, $\bm g_t \in \mathbb{R}^{D}$. 
Traditional TTA methods conduct adaptation independently at each time step, which can exacerbate the issue of cumulative errors~\cite{niu2022efficient,song2023ecotta}, particularly in the VLN process that requires frequent action execution.
Therefore, we propose to conduct a gradient decomposition-accumulation analysis, wherein we periodically analyze the gradients generated during the recent multi-step navigation process and identify a concordant direction for an iteration of model update.

\Paragraph{Gradient Decomposition-Accumulation.}
During navigation, 
as shown in~\figref{framework}, we perform model update every $M$ action steps. For the $j$-th update, gradients from previous $M$ steps are collected as $\bm G_j= \{\widetilde{\bm g}_{j,m}\}_{m=1}^{M}$, where $\bm G_j \in \mathbb{R}^{M \times D}$, $\widetilde{\bm g}_{j,m}$ indicates the $t$-th gradient $\bm g_{t}$ when $t=M(j-1)+m$. Note that these gradients determine the learning direction of our VLN model, and a simple strategy to compute this direction is to take their average $\bar{\bm g}_j = 1/M\sum_m \widetilde{\bm g}_{j,m}$; however, this inevitably introduce step-specific noise. To avoid the issue, we aim to find a concordant direction 
among these gradients. We first establish a local 
coordinate system with $D$ orthogonal and unit axes (bases) $\bm U_j = \{{\bm u_{j,d}}^\mathsf{T} \}_{d=1}^D \in  \mathbb{R}^{D\times D}$ for gradient decomposition, where each gradient can be approximatively linearly represented by these bases. Intuitively, the axes along which the gradients exhibit the higher variance after projection represent the directions of gradients with the lower consistency. These directions have the potential to introduce interference in determining a model update direction. Therefore, it is advisable to reduce the projection of gradients in these directions. To solve the bases $\bm U_j$, we  can
utilize singular value decomposition (SVD) as follows:

\begin{equation}\label{eq:svd}
	\lambda_{j,d}, \bm u_{j,d} =  \textbf{SVD}_d\left(\frac{1}{M-1}{\hat{\bm G}}_j^\mathsf{T}\hat{\bm G}_j\right),
\end{equation}
where $\hat{\bm G}_j$ is the centered gradient matrix by removing the mean from $\bm G_j$. The $m$-th row 
in $\hat{\bm G}_j$ reflects the deviation between $\widetilde{\bm g}_{j,m}$ and the average gradient $\bar{\bm g}_j$. $\lambda_{j,d}$, $\bm u_{j,d}$ denote the $d$-th largest eigenvalue and the corresponding eigenvector. Motivated by the principle component analysis~\cite{shlens2014tutorial}, it is obvious that a larger $\lambda_{j,d}$ corresponds to a higher variance of the gradient projection length $\bm G_j \bm u_{j,d}$ and vice versa. Hence, we can derive a concordant gradient by adaptively aggregating the gradients' components on all the axes by considering different eigenvalue (importance):

\begin{equation}\label{eq:gradient_fast}
	\nabla_{j}^{(fast)} = \sum\nolimits_{d=1}^{D} \Phi_{d}(\lambda_{j,d}) \cdot <\bar{\bm g}_j,\bm u_{j,d}>\bm u_{j,d},
\end{equation}
where the last term denotes the projected component of the averaged gradient $\bar{\bm g}_j$ on to the $d$-th axis. $\Phi_{d}(\cdot)$ is referred as the adaptive coefficient for accumulating all the components, which is simply defined as $\Phi_{d}(\lambda_{j,d}) = 1/\lambda_{j,d}$, reflecting the importance of various axes. Notably, when removing the coefficient, $\nabla_{j}^{(fast)}$ is degenerated into $\bar{\bm g}_j$, which is used in regular gradient descent approaches.

Based on~\eqnref{gradient_fast}, a concordant optimization direction is established by enhancing the components that are convergent among $\{\widetilde{\bm g}_{j,m}\}_{m=1}^{M}$ and suppressing those divergent ones. However, the introduction of $\Phi_{d}(\cdot)$ makes the length of $\nabla_{j}^{(fast)}$ uncontrollable. Therefore, we calibrate its length to $\left \| \bar{\bm g}_j \right \|_2$, which encodes the gradient length from the last three time steps, for a more reasonable model update:
\begin{equation}\label{eq:gradient_calib}
	\nabla_{j}^{(fast)} \gets  ( \nabla_{j}^{(fast)} \| \bar{\bm g}_j  \|_2) /  \| \nabla_{j}^{(fast)}  \|_2
\end{equation}

With $\nabla^{(fast)}$, we can perform fast model update by setting a learning rate $\gamma^{(fast)}$. Although traditional methods employ a fixed learning rate, 
such a setting might hinder model convergence, \emph{i.e.}, small learning rates slow down convergence while aggressive learning rates prohibit convergence~\cite{Barzilai1988TwoPointSS}. Since fast updates are frequently invoked during navigation, relying on a fixed learning rate is sub-optimal. Therefore, we propose to dynamically adjust learning rate throughout the fast update phase.



\Paragraph{Dynamic Learning Rate Scaling.} 
Different from varying the learning rate through optimizer or scheduler, we argue for a scaling method that leverages gradient agreement information in historical steps to dynamically adjust the speed of model update.
Current gradient alignment strategies typically impose direct constraints on the gradients~\cite{shi2021gradient,rame2022fishr}, which are not suitable for our framework as they undermine the gradient decomposition-accumulation process. 
Given that the second-order information (variance) has been demonstrated  to be more effective than the first-order information (mean) in gradient agreement learning~\cite{rame2022fishr}, we directly utilize the trace of the gradient covariance matrix, $\text{Tr}\left(1/(M-1){\hat{\bm G}}_j^\mathsf{T}\hat{\bm G}_j\right)$, for scaling. Note that the trace is equal to the sum of eigenvalues $\sigma_j = \sum_d \lambda_{j,d}$. Here, when $\sigma_j$ deviates significantly from the historical variance, we assign a smaller learning rate, and vice versa:

\begin{equation}\label{eq:lr}
	\gamma^{(fast)}_j = \text{Trunc}\left(1 + \tau - |\sigma_j- \bar{\sigma}|\right) \cdot \hat{\gamma}^{(fast)},
\end{equation}
where $\text{Trunc}(\cdot)$ is the truncation function that truncates the input to the interval $[a, b]$. $\tau$ is a threshold and $\hat{\gamma}^{(fast)}$ is the base learning rate. The historical variance $\bar{\sigma}$ is updated as $\bar{\sigma} \gets \rho\bar{\sigma} + (1-\rho)\sigma_j$ and maintained for all samples throughout the test stage, $\rho$ is the update momentum.  



\Paragraph{Model Update.}
With the above gradient and learning rate, we can perform a single iteration as the $j$-th fast update:
\begin{equation}\label{eq:fast}
	\bm{\Theta}_{j} = \bm{\Theta}_{j-1} - \gamma_{j}^{(fast)} \cdot \nabla_{j}^{(fast)},
\end{equation}
where the subscript of $\bm{\Theta}$ indicates the index of model update in the current test sample.

\vspace{\subsecmargin}
\subsection{Slow Update via Parameter Analysis}
\vspace{\subsecmargin}
In the fast update phase, although we  obtain concordant optimization directions, the frequent parameter updates may still dramatically change the VLN model. 
To maintain the stability of the VLN model during online long-term usage,
we periodically revert the model to its historical states, 
and conduct a decomposition-accumulation analysis on the parameter variation trajectory for direct parameter modulation. The slow update phase shares the core formulation 
with the fast phase, but shifts the focus from gradients to the model parameters themselves.

\Paragraph{Parameter Decomposition-Accumulation.}
Following the completion of the fast update phase on the $o$-th test sample, the model state (parameters) is recorded as $\bm{\Theta}_{o,J_o}$, where $J_o$ denotes the final fast update step on this sample, and the subscript $o$ has been omitted in the previous section. We then treat these historical states as a parameter variation trajectory to facilitate stable model updates. As shown in the right part of~\figref{framework}, the slow model update is invoked every $N$ samples. For the $l$-th update, historical model states are collected as $\bm M_l = \{\widetilde{\bm{\Theta}}_{l,n}\}_{n=0}^N$, where $\bm M_l \in \mathbb{R}^{(N+1) \times D}$, $\widetilde{\bm{\Theta}}_{l,n}$ indicates 
the $o$-th 
model state $\bm{\Theta}_{o,J_o}$ 
when $o=N(l-1)+n$ and $n \neq 0$.  $\widetilde{\bm{\Theta}}_{l,0}$ indicates the model state produced by the previous slow update, and we use it interchangeably with $\bm{\Theta}^{(l-1)}$ in the following. Note that in the slow update phase, we additionally incorporate $\bm{\Theta}^{(l-1)}$  from the previous update for analysis since it serves as a starting reference point for direct parameter modulation.

\begin{table*} \centering
	\caption{Experimental results for different TTA strategies on REVERIE dataset.}
	\label{tab:tta}
	\resizebox{\textwidth}{!}{
		\large
		\begin{threeparttable}
		\begin{tabular}{l|*{4}{c}|*{4}{c}|*{4}{c}|c}
			\toprule
			\multirow{2}{*}{\textbf{Methods}}
			& \multicolumn{4}{c}{\textbf{REVERIE} Val Seen}
			& \multicolumn{4}{|c}{\textbf{REVERIE} Val Unseen} 
			& \multicolumn{4}{|c|}{\textbf{REVERIE} Test Unseen} 
			& \multirow{2}{*}{\textbf{Time(ms)}\tnote{*}}\\
			& \cellcolor{blue!25}OSR & \cellcolor{blue!25}SR & \cellcolor{blue!25} SPL & \cellcolor{blue!25}RGSPL
			& \cellcolor{blue!25}OSR & \cellcolor{blue!25}SR & \cellcolor{blue!25} SPL & \cellcolor{blue!25}RGSPL
			& \cellcolor{blue!25}OSR & \cellcolor{blue!25}SR & \cellcolor{blue!25} SPL & \cellcolor{blue!25}RGSPL  \\
			\midrule
			DUET~\cite{chen2022think} 
			& 73.86 & 71.75 & 63.94 & 51.14
			& 51.07 & 46.98 & 33.73 & 23.03  
			& 56.91 & 52.51 & 36.06 & 22.06 & 104.84\\
			\midrule
			\quad+ EATA~\cite{niu2022efficient} 
			& 74.05 & 72.17 & 63.41 & 49.94      
			& 52.09 & 47.40 & 33.46 & 22.65     
			& 57.21 & 52.91 & 35.19 & 21.65 & 133.12\\
			\quad+ CoTTA~\cite{wang2022continual} 
			& 73.79 & 71.05 & 62.91 & 49.36			
			& 52.46 & 47.56 & 31.43 & 21.83 
			& 56.14 & 52.52 & 34.66 & 21.06 & 3.89$\times$ 10$^{3}$\\
			\quad+ NOTE~\cite{gong2022note} 
			& 74.76 & 72.43 & 64.28 & 51.97
			& 52.85 & 48.28 & 33.98 & 22.98 
			& 57.66 & \textbf{53.41} & 36.18 & 22.09 & 137.89\\
			\quad+ SAR~\cite{niu2023towards} 
			& 74.84 & 71.75 & 64.43 & 51.70
			& 53.26 & 48.00 & 33.92 & 23.09 
			& 57.11 & 53.04 & 36.07 & 22.27 & 145.53\\
			\quad+ ViDA~\cite{Liu2023ViDAHV} 
			& 73.99 & 72.49 & 63.49 & 50.89
			& 52.53 & 48.14 & 32.45 & 21.92 
			& 56.78 & 52.74 & 35.10 & 21.77 & 3.97$\times$ 10$^{3}$\\
			\midrule
			\quad+ Tent~\cite{wang2020tent} 
			& 70.07 & 70.73 & 61.67 & 49.31
			& 49.43 & 46.87 & 31.90 & 20.15 
			& 54.58 & 50.56 & 33.37 & 20.32 & 126.91\\
			\quad+ Tent-INT-2 
			& 71.82 & 70.36 & 61.53 & 49.82
		    & 51.22 & 48.46 & 33.67 & 21.30  
			& 54.25 & 50.36 & 33.89 & 21.09 & 124.02\\
			\quad+ Tent-INT-3 
			& 74.54 & 72.58 & 64.44 & 51.05
		    & 52.28 & 48.60 & 34.65 & 23.12  
			& 56.66 & 52.92 & 35.84 & 21.89 & 119.34\\
			\quad+ Tent-INT-4 
			& 73.84 & 72.51 & 64.03 & 50.97 
			& 51.40 & 48.91 & 35.06 & 22.99  
			& 56.74 & 53.14 & 36.29 & 22.14 & 117.26\\
			\quad+ Tent-Stable 
			& 73.72 & 71.89 & 64.06 & 50.41
			& 51.43 & 47.55 & 33.99 & 23.32
			& 57.12 & 52.61 & 36.17 & 22.16 & 129.22\\
			\midrule
			\quad+ FSTTA 
			& \textbf{75.59} & \textbf{75.48} & \textbf{65.84} & \textbf{52.23} 
			& \textbf{56.26} & \textbf{54.15} & \textbf{36.41} & \textbf{23.56} 
			& \textbf{58.44} & 53.40 & \textbf{36.43} & \textbf{22.40} & 135.61\\
			\bottomrule
		\end{tabular}
	    \begin{tablenotes}
	    	\footnotesize
	    	\item[*] Note that the last column displays the average execution time of the agent for a single instruction, calculated on the Validation Unseen set of REVERIE.
	    \end{tablenotes}
	\end{threeparttable}
	}
\end{table*}

Similar to the fast update phase, the centered parameter matrix $\hat{\bm M}_l$ can be constructed, where the $n$-th row vector in it reflects the deviation between $\widetilde{\bm{\Theta}}_{l,n}$ and the averaged historical parameter $\bar{\bm{\Theta}}_{l} = 1/(N+1)\sum_n\widetilde{\bm{\Theta}}_{l,n}$. 
With $\hat{\bm M}_l$, we can obtain the following eigenvalues and eigenvectors: $\epsilon_{l,d}, \bm z_{l,d} = \textbf{SVD}_d(1/N \cdot {\hat{\bm M}}_l^\mathsf{T}\hat{\bm M}_l)$, where a larger $\epsilon_{l,d}$ corresponds to a higher variance of the parameter projection length ${\bm M}_l\bm z_{l,d}$ and vice versa. $\bm Z_l = \{{\bm z_{l,d}}^\mathsf{T}\}_{d=1}^D$ depicts the local 
coordinate system where each axis depicts the direction of parameter variation. Intuitively, the principal axes (with larger eigenvalues) delineate the primary directions of historical parameter variation, while minor axes (with smaller eigenvalues) often encompass noise~\cite{wang2022pgrad}. 
To find a reliable optimization path to traverse the trajectory of primary parameter changes, we pay attention on the axis with the large variance.
Since there is no silver bullet to learning an 
optimization direction with only parameters, a reference direction can significantly aid in guiding the model towards a local optimal. Here, we leverage the parameter variations to calculate the reference direction:

\begin{equation}\label{eq:re}
	\bm h_l=\frac{1}{\sum_{i=0}^{N-1} q^{i}} \sum\nolimits_{n=1}^{N} q^{N-n} \cdot (\widetilde{\bm{\Theta}}_{l,0}-\widetilde{\bm{\Theta}}_{l,n}),
\end{equation}
where the hyper-parameter $q \in (0,1)$, which assigns larger weight to the more recent parameter deviations as they encapsulate richer sample information. Then, we calculate the optimization path (gradient) in the slow update phase as: 

\begin{equation}\label{eq:gradient_slow}
	\nabla_{l}^{(slow)} = \sum\nolimits_{d} \Psi_{d}(\bm \epsilon_l, \bm h_l) \cdot \text{sign} \left(<\bm h_l, \bm z_{l,d}>\right) \bm z_{l,d},
\end{equation}
where the use of sign function $\text{sign}(\cdot)$ is to force the axes to be positively related to the reference direction $\bm h_l$. Notably, different from~\eqnref{gradient_fast} that uses the projected components on each axis for estimating an optimization direction, here we only utilize the axes themselves for deriving $\nabla_{l}^{(slow)}$. The reason is that these axes depict the parameter variation direction, which can be directly used for estimating gradients.  $\Psi_{d}(\cdot)$ is referred as the adaptive coefficient for accumulating all the axes (optimization directions), defined as:

\begin{equation}\label{eq:phi}
	\Psi_{d}(\bm \epsilon_l, \bm h_l) =  \frac{\epsilon_{l,d} \cdot \| \bm h_l \|_2}{ \| \bm \epsilon_l \|_2},
\end{equation}
where the L2-normalization is performed on eigenvalues   to convey different relative importance of axes. Besides, the norm of the reference direction is utilized to automatically tuning the magnitude of the analyzed gradient. In contrast to  $\Phi_{d}(\cdot)$ in the fast update phase, $\Psi_{d}(\cdot)$ highlights those axes with high variation due to the different characteristics of gradients and parameters in model optimization. 

%

\Paragraph{Model Update.}
With $\nabla_l^{(slow)}$, we can perform the $l$-th slow model update as follows:
\begin{equation}\label{eq:fast}
	\bm{\Theta}^{(l)} = \bm{\Theta}^{(l-1)} - \gamma^{(slow)} \cdot \nabla_{l}^{(slow)},
\end{equation}
where $\gamma^{(slow)}$ is learning rate. Since the slow update phase is designed for stable model learning and is not frequently invoked, we employ a fixed learning rate here instead of conducting the dynamic learning rate scaling as done in the fast phase. The updated parameter $\bm{\Theta}^{(l)}$ will be utilized for the subsequently coming test samples in conjunction with new fast update phases applied to them. 
\section{Experimental Results}
\label{sec:expr}



\Paragraph{Datasets.} 
We use the popular and standard VLN benchmark REVERIE~\cite{qi2020reverie} to investigate test-time adaptation within the realm of online VLN. REVERIE contains 10,567 panoramic images and 21,702 high-level instructions, focusing on grounding remote target object within 90 buildings. In addition, we also adopt other three benchmarks for evaluating the effectiveness of our proposed FSTTA. Among them, R2R~\cite{anderson2018vision} provides step-by-step instructions for navigation in photo-realistic environments, which includes 10,800 panoramic views and 7,189 trajectories. SOON~\cite{zhu2021soon} also requires the agent to find the target object with a more detailed description of the goal. It has 3,848 sets of instruction and more than 30K long distance trajectories. 
Note that the performance comparisons are based on \href{https://scenario-oriented-objectnavigation.github.io}{their challenge report}, specifically within the provided val unseen and test unseen splits.
R2R-CE~\cite{krantz2020beyond} is a variant of R2R in continuous environments, where an agent is able to move freely and engage with obstacles. The dataset consists of 16,000 instruction-trajectory pairs, with non-transferrable paths excluded.


\Paragraph{Evaluation Metrics.}
We follow previous approaches~\cite{qi2020reverie,chen2022think,chen2022learning,li2022envedit,wang2023gridmm} and employ the most commonly used metrics for evaluating VLN agents, i.e., TL (Trajectory Length): the agent's average path length in meters; NE (Navigation Error): average distance in meters between the agent's final location and the target one; SR (Success Rate): the proportion of successfully executed instructions with the NE less than 3 meters; SPL (Success weighted by Path Length): SR penalized by Path Length, which is calculated as $\frac{1}{E} \sum_{i=1}^{E} S_i \frac{l_i}{max(p_i,l_i)}$, where E is the number of tasks, $S_i$ denotes the success as a binary value, $l_i$ and $p_i$ denote the shortest path and actual path length for the $i^{th}$ task; OSR (Oracle Success Rate): SR given the oracle stop policy; RGS (Remote Grounding Success rate): proportion of successfully executed instructions where the output bounding box has an IoU (intersection over union) $\ge$ 0.5 with the ground truth; and RGSPL (RGS weighted by Path Length): RGS penalized by Path Length.
Among them, SR and SPL are the most common metrics for evaluation.
Note that only the optimal values of experimental results are highlighted in bold across all tables. Moreover, from Tables~\ref{tab:reverie} to~\ref{tab:r2rce}, different font colors are employed to indicate whether our method exceeds the performance of the corresponding baseline methods, \emph{i.e.}, using red to denote superior results and blue for inferior ones.


\Paragraph{Implementation Details.}
To better conform to 
practical applications, 
we set batch size to 1 during evaluation, where each sample (and each action step) is forward propagated only once. 
Owing to the lack of an authentic online VLN evaluation setting, we shuffle test samples in each dataset split and sequentially input them into the agent to simulate the online execution and adaptation.
Specifically, for VLN models equipped with TTA strategies, we run the experiments with shuffled samples 5 times and report the average results.
We adopt DUET~\cite{chen2022think} and HM3D~\cite{chen2022learning} as the base models. Since HM3D does not provide training code for R2R-CE, we adopt another SOTA methods, WS-MGMap~\cite{Chen2022WeaklySupervisedMM} and BEVBert~\cite{an2023bevbert}, for TTA. 
Note that for the base models, 
we report the results obtained from running their official codes.
In our FSTTA, we only utilize the last four LN layers of base models for model updating, all the feature dimensions of these layers are 768. We set the intervals for fast and slow updates to $M=3$ and $N=4$, the learning rates of the two phases are $\hat{\gamma}^{(fast)}=6 \times 10^{-4}$ and $\gamma^{(slow)}=1 \times 10^{-3}$. For the dynamic learning rate scaling, we empirically set the threshold $\tau=0.7$ in~\eqnref{lr} and the update momentum $\rho=0.95$ with the truncation interval $[0.9,1.1]$. And the hyper-parameter $q$ in \eqnref{re} is set to 0.1.
All experiments are conducted on a RTX 3090 GPU. 


\begin{table}
	\centering
	\caption{Ablation study on REVERIE dataset.}
	\label{tab:ablation}
	\resizebox{0.48\textwidth}{!}{
		\large
		\begin{tabular}{*{3}{c}|*{6}{c}}
			\toprule
			\multicolumn{3}{c}{\textbf{Module}} & \multicolumn{6}{c}{\textbf{REVERIE} Val Unseen } \\ 
			\midrule
			Fast & DLR & Slow & \cellcolor{red!25}TL $\downarrow$  & \cellcolor{blue!25}OSR & \cellcolor{blue!25}SR & \cellcolor{blue!25} SPL &
			\cellcolor{blue!25}RGS & \cellcolor{blue!25}RGSPL\\
			\midrule
			- &  - & -  & \textbf{22.11} & 51.07 & 46.98 & 33.73 & 32.15 & 23.03 \\
			\midrule
			Tent&  - &  - & 22.52 & 52.28 & 48.60  & 34.65  & 32.66  & 23.12  \\
			\midrule
			\checkmark & -  & -  & 22.65 & 53.50 & 49.74 & 34.91 & 33.70 & 23.36 \\
			\checkmark &\checkmark & -  & 22.43 & 54.01 & 49.82 & 35.34 & \textbf{34.32} & 23.29 \\
			\checkmark &\checkmark &\checkmark & 22.14 & \textbf{56.26} & \textbf{54.15} & \textbf{36.41} & 34.27 & \textbf{23.56} \\
			\bottomrule
		\end{tabular}
	}
\end{table}

\begin{table}
	\centering
	\caption{Results on Validation Seen set of REVERIE.}
	\label{tab:forgetting}
	\resizebox{0.48\textwidth}{!}{
		\large
		\begin{tabular}{*{2}{c}|*{6}{c}}
			\toprule
			\multicolumn{2}{c}{FSTTA} & \multicolumn{6}{c}{\textbf{REVERIE} Val Seen } \\ 
			\midrule
			Unseen &Seen & \cellcolor{red!25}TL $\downarrow$  & \cellcolor{blue!25}OSR & \cellcolor{blue!25}SR & \cellcolor{blue!25} SPL &
			\cellcolor{blue!25}RGS & \cellcolor{blue!25}RGSPL\\
			\midrule
			\multicolumn{2}{c|}{\ $\times$ \ ~~~~~~~ \ $\times$}& 13.86  & 73.86 & 71.15 & 63.94 & 57.41 & 51.14 \\
			\midrule
			\multicolumn{2}{c|}{\ $\times$ \ $\longrightarrow$ \ \checkmark}& 15.13 & \textbf{75.59} & \textbf{75.48} & \textbf{65.84} & 58.62 & \textbf{52.23} \\
			\multicolumn{2}{c|}{\ \checkmark \ \ $\longrightarrow$  \ $\times$} & \textbf{13.40} & 73.16 & 71.78 & 64.18 & 57.05 & 51.18 \\
			\multicolumn{2}{c|}{\ \checkmark \ \ $\longrightarrow$  \ \checkmark}& 15.11 & 75.58 & 74.12 & 65.53 & \textbf{59.20} & 52.18 \\
			\bottomrule
		\end{tabular}
	}
\end{table}

\begin{table}
	\centering
	\caption{Results on Validation Unseen \& Seen sets of REVERIE.}
	\label{tab:generalized}
	\resizebox{0.48\textwidth}{!}{
		\large
		\begin{tabular}{l|*{6}{c}}
			\toprule
			\multirow{2}{*}{\textbf{Methods}}
			& \multicolumn{6}{c}{\textbf{REVERIE} Val Unseen \& Seen} \\
			&\cellcolor{red!25}TL $\downarrow$  & \cellcolor{blue!25}OSR & \cellcolor{blue!25}SR & \cellcolor{blue!25} SPL &
			\cellcolor{blue!25}RGS & \cellcolor{blue!25}RGSPL \\ 
			\midrule
			DUET 
			& \textbf{19.18}  & 61.53  & 57.49  & 45.66  & 41.56  & 34.38   \\
			\midrule
			\quad+ Tent 
			& 20.23  & 57.33  & 54.86  & 41.90  & 38.09  &  32.46  \\
			\quad+ EATA 
			& 20.29  & 62.77  & 57.31  & 44.59  & 41.54  &  34.16  \\
			\quad+ SAR 
			&  20.52 & \textbf{63.59}  & 57.80  & 44.72  &  42.45 &  34.88  \\
			\midrule
			\quad+ FSTTA & 20.48 & 63.36 & \textbf{60.23} & \textbf{47.96} & \textbf{43.58} & \textbf{35.65} \\
			\bottomrule
		\end{tabular}
	}
\end{table}

\vspace{\subsecmargin}
\subsection{Comparison with Different TTA Strategies}
\label{sec:tta_compar}
\vspace{\subsecmargin}
Currently, various TTA methods have been adeptly integrated for the dynamic model updates, 
marking significant progress. Although the exploration of TTA
within the VLN field remains relatively untapped, the integration of contemporary advanced TTA methodologies into VLN 
is feasible. 
Since efficiency is an important evaluation metric for TTA, we provide the average time taken by each method to execute a single instruction for comparison.
For the compared methods, SAR and TENT are the popular entropy minimization models, whereas NOTE, CoTTA, EATA and ViDA are state-of-the-art continual TTA methods. The results  in~\tabref{tta} demonstrate the capability of our proposed FSTTA to blend model performance with testing efficiency. Specifically, on the validation unseen dataset,  our method exhibits a discernible enhancement of 6.2\% and 2.5\% on the SR and SPL metrics compared to the state-of-the-art SAR method, concurrently manifesting a reduction of 7\% in testing time. 
From the results, we observe that directly applying existing TTA methods to the online VLN task does not lead to significant performance improvements. Furthermore, we investigate different frequencies of updates based on TENT as well as the stable update approach. `INT' denotes the update interval, which means averaging the gradient information over a certain interval and then performing an iteration of model update; these results are consistent with those in~\figref{motivation}(b). It can be seen that our method still outperforms these strategies with marginally increased time costs. 



\begin{table*} 
	\centering
	\caption{Experimental results on REVERIE dataset.}
\label{tab:reverie}
\resizebox{0.98\textwidth}{!}{
	\begin{tabular}{l|*{4}{c}|*{4}{c}|*{4}{c}}
		\toprule
		\multirow{2}{*}{\textbf{Methods}}
		& \multicolumn{4}{c}{ Val Seen} 
		& \multicolumn{4}{|c}{ Val Unseen} 
		& \multicolumn{4}{|c}{ Test Unseen} \\
		& \cellcolor{blue!25}OSR & \cellcolor{blue!25}SR & \cellcolor{blue!25} SPL & \cellcolor{blue!25}RGSPL 
		& \cellcolor{blue!25}OSR & \cellcolor{blue!25}SR & \cellcolor{blue!25} SPL & \cellcolor{blue!25}RGSPL 
		& \cellcolor{blue!25}OSR & \cellcolor{blue!25}SR & \cellcolor{blue!25} SPL & \cellcolor{blue!25}RGSPL \\
		\midrule
		Seq2Seq~\cite{anderson2018vision} 
		& 35.70 & 29.59  & 24.01  & 14.96
		& 8.07 & 4.20   & 2.84 & 2.16  
		& 6.88 & 3.99  & 3.09 & 1.58\\
		RCM~\cite{wang2019reinforced} 
		& 29.44 & 23.33  & 21.82 & 15.36  
		& 14.23 & 9.29  & 6.97  & 3.89
		& 11.68 & 7.84 & 6.67 & 3.14  \\
		FAST~\cite{qi2020reverie} 
		& 55.17 & 50.53 & 45.50 & 29.66
		& 28.20 & 14.40 & 7.19 & 4.67
		& 30.63 & 19.88  & 11.61  & 6.08\\
		SIA~\cite{Lin2021SceneIntuitiveAF} 
		& 65.85 & 61.91 & 57.08 & 42.65 	
		& 44.67 & 31.53 & 16.28  & 11.56 
		& 44.56 & 30.80 & 14.85 & 9.20\\
		RecBERT~\cite{hong2021vln} 
		& 53.90 & 51.79 & 47.96 & 35.61
		& 35.02 & 30.67 & 24.90 & 15.27
		& 32.91 & 29.61 & 23.99 & 13.51\\
		Airbert~\cite{Guhur2021AirbertIP} 
		& 48.98 & 47.01 & 42.34 & 30.01
		& 34.51 & 27.89 & 21.88 & 14.18
		& 34.20 & 30.28 & 23.61 & 13.28 \\
		HAMT~\cite{chen2021history}
		& 47.65 & 43.29 & 40.19 & 25.18
		& 36.84 & 32.95 & 30.20 & 17.28
		& 33.41 & 30.40 & 26.67 & 13.08\\
		HOP~\cite{Qiao2022HOPHA} 
		& 53.76 & 54.88 & 47.19 & 33.85
		& 36.24 & 31.78 & 26.11 & 15.73
		& 33.06 & 30.17  & 24.34& 14.34\\
		BEVBert~\cite{an2023bevbert}
		& \textbf{76.18} & 73.72 & 65.32 & 51.73
		& 56.40 & 51.78 & 36.37 & 24.44
		& 57.26 & 52.81 & 36.41 & 22.09\\
		LANA~\cite{Wang2023LANAAL}
		& 74.28 & 71.94 & 62.77 & 50.34 
		& 52.97 & 48.31 & 33.86 & 22.77
		& 57.20 & 51.72 & 36.45 & 22.85\\
		GridMM~\cite{wang2023gridmm} 
		& - & - & - & -
		& 57.48 & 51.37 & 36.47 & 24.56
		& 59.55 & 55.13 & 36.60 & \textbf{23.45}\\
		\midrule
		DUET~\cite{chen2022think} 
		& 73.86 & 71.75 & 63.94 & 51.14 
		& 51.07 & 46.98 & 33.73 & 23.03
		& 56.91 & 52.51 & 36.06 & 22.06\\
		DUET-FSTTA 
		& \textcolor{red}{75.59} & \textcolor{red}{\textbf{75.48}} & \textcolor{red}{\textbf{65.84}} & \textcolor{red}{\textbf{52.23}}
		& \textcolor{red}{56.26} & \textcolor{red}{54.15}  & \textcolor{red}{36.41}  & \textcolor{red}{23.56}
		& \textcolor{red}{58.44} & \textcolor{red}{53.40}  & \textcolor{red}{36.43}  & \textcolor{red}{22.40}\\
		\midrule
		HM3D~\cite{chen2022learning} 
		& 66.76 & 65.00 & 55.70 & 41.66
		& 62.11 & 55.89 & 40.85 & \textbf{26.76} 
		& 59.81 & 53.13 & 38.24 & 22.68\\
		HM3D-FSTTA
		& \textcolor{red}{69.41} &\textcolor{red}{67.79}  & \textcolor{red}{58.28}  & \textcolor{blue}{41.60}
		& \textcolor{red}{\textbf{63.74}} & \textcolor{red}{\textbf{57.02}} & \textcolor{red}{\textbf{41.41}} & \textcolor{blue}{26.55}
		& \textcolor{red}{\textbf{63.68}} & \textcolor{red}{\textbf{56.44}}  & \textcolor{red}{\textbf{39.58}} & \textcolor{red}{23.04} \\
		\bottomrule
	\end{tabular}
}
\end{table*}

\begin{table} [t!]
	\centering
	\caption{Experimental results on R2R dataset. }
	\label{tab:r2r}
	\resizebox{\columnwidth}{!}{
		\begin{tabular}{l|*{4}{c}|*{4}{c}}			
			\toprule
			\multirow{2}{*}{\textbf{Methods}} & \multicolumn{4}{c}{ Val Seen}& \multicolumn{4}{|c}{ Val Unseen} \\
			&\cellcolor{red!25}TL $\downarrow$  & \cellcolor{red!25}NE $\downarrow$
			& \cellcolor{blue!25}SR & \cellcolor{blue!25} SPL  
			&\cellcolor{red!25}TL $\downarrow$  & \cellcolor{red!25}NE $\downarrow$
			& \cellcolor{blue!25}SR & \cellcolor{blue!25} SPL  	\\
			\midrule
			Seq2Seq~\cite{anderson2018vision} 
			& 11.33 & 6.01 & 39 &-
			& \textbf{8.39} & 7.81 & 22 &- \\
			RCM~\cite{wang2019reinforced} 
			& 10.65 & 3.53 & 67 & -
			& 11.46 & 6.09 & 43 & - \\
			EnvDrop~\cite{tan2019learning} 
			& 11.00 & 3.99 & 62 & 59 
			& 10.70 & 5.22 & 52 & 48 \\
			PREVALENT~\cite{hao2020towards} 
			& \textbf{10.32} & 3.67 & 69 & 65
			& 10.19 & 4.71 & 58 & 53 \\
			RecBERT~\cite{hong2021vln} 
			& 11.13 & 2.90 & 72 & 68
			& 12.01 & 3.93 & 63 & 57\\
			HAMT~\cite{chen2021history} 
			& 11.15 & 2.51 & 76 & 72
			& 11.46 & 3.65 & 66 & 61\\
			HOP~\cite{Qiao2022HOPHA} 
			& 11.26 & 2.72 & 75 & 70
			& 12.27 & 3.80 & 64 & 57\\
			DAVIS~\cite{lu2022anticipating}
			& 12.45 & 3.16 & 80 & \textbf{76}
			& 12.65 & 3.16 & 67 & 61\\
			BEVBert~\cite{an2023bevbert}
			& 13.56 & \textbf{2.17} & \textbf{81} & 74
			& 14.55 & 2.81 & \textbf{75} & \textbf{64} \\
			\midrule
			DUET~\cite{chen2022think}
			& 12.33 & 2.28 & 79 & 73
			& 13.94 & 3.31 & 72 & 60 \\
			DUET-FSTTA 
			& \textcolor{blue}{13.39} & \textcolor{red}{2.25} & \textcolor{red}{79} & \textcolor{red}{73}
			& \textcolor{blue}{14.64} & \textcolor{red}{3.03} & \textcolor{red}{\textbf{75}} & \textcolor{red}{62} \\
			\midrule
			HM3D~\cite{chen2022learning} 
			& 13.30 & 2.70 & 77 & 71
			& 14.29 & 2.83  & 74  & 62  \\
			HM3D-FSTTA
			& \textcolor{blue}{13.52} & \textcolor{red}{2.57} & \textcolor{red}{78} & \textcolor{red}{71}
			& \textcolor{blue}{14.86} & \textcolor{red}{\textbf{2.71}}  & \textcolor{red}{\textbf{75}}  & \textcolor{red}{63}  \\		
			\bottomrule
		\end{tabular}
	}
\end{table}

\vspace{\subsecmargin}
\subsection{Extensive Analysis of FSTTA}
\vspace{\subsecmargin}

\Paragraph{Ablation Studies of the Proposed FSTTA.}
In this work, we propose a FSTTA method for online VLN, which consists of both fast and slow model update phases. To validate their effectiveness, we progressively integrate the two phases into the baseline DUET model. In addition, 
we design a baseline variant, 
which equips DUET with the vanilla TTA objective (TENT~\cite{wang2020tent}) and simply utilize the averaged gradient in an interval (with the same $M$) for fast model updates.
Empirical findings from~\tabref{ablation} illuminate that the integration of fast and slow phases progressively bolsters the base model  by 2.8\% and 4.3\% on the SR metric. Moreover, the dynamic learning rate scaling module (DLR) also contributes to enhancing the model's performance.  

\Paragraph{Will our method experience catastrophic forgetting?}
For an online VLN agent endowed with the TTA capability, it faces the issue of catastrophic forgetting of historical environments and instructions upon continually executing new instructions in new environments. To assess whether our method harbors this issue, we re-evaluate our methods on the REVERIE \emph{validation seen} set. 
Compared with the base model, as shown in~\tabref{forgetting}, we find that: (1) Obviously, directly applying FSTTA with the base model on seen data can noticeably enhance performance. (2) After performing FSTTA on the unseen set, the obtained model, when tested directly on the seen dataset without TTA, achieves performance comparable to the base model, confirming that our method does not suffer from catastrophic forgetting. (3) Applying the updated model from unseen set to the seen set with TTA yields comparable results against the seen-only-TTA version. This indicates that our method is effective in environment adaption and experience accumulation.

\Paragraph{Generalization Testing in More Practical Environments.}
In practical applications, agents might encounter both previously seen and unseen scenarios. In preceding experiments, we exclusively test on the validation seen and unseen sets separately. To verify the generalizability, we combine the seen and unseen sets into a unified set for online VLN.~\tabref{generalized} shows that FSTTA outperforms other TTA methods in effectively managing a variety of testing scenarios.

\vspace{\subsecmargin}
\subsection{Comparison with State-of-the-art VLN Models}
\label{sec:sota_comparison}
\vspace{\subsecmargin}
\Paragraph{REVERIE.}
\tabref{reverie} presents a comparison on REVERIE dataset. 
Compared with the base models which do not perform test-time adaptation, the proposed method shows favorable performance improvement across most evaluation metrics across the three dataset splits. Specifically, on the validation unseen split, our model exhibits notable advantages over DUET, with improvements of 7.1\% on SR, and 2.7\% on SPL. 
These results unequivocally affirm the effectiveness of our fast-slow test time adaptation model, showing the promising potential of TTA in the VLN field. 

\Paragraph{R2R.}
\tabref{r2r} shows the comparison results on R2R dataset. 
Our approach outperforms the base models in most metrics 
(\emph{e.g.}, $72\% \rightarrow 75\%$ for DUET on SR, $62\% \rightarrow 63\%$ for HM3D on SPL).
Notably, from the results of the above two datasets, our method, while enhancing the success rate of VLN, causes a slight increase in the path length (TL). We speculate that a possible reason is that performing TTA online may increase the likelihood of the agent deviating from its original action execution pattern, leading to more exploration or backtracking. This situation is also confirmed in the analysis of various TTA strategies in~\tabref{tta}.

\Paragraph{SOON.}
The proposed FSTTA establishes new state-of-the-art results across most metrics on this dataset. For instance, as shown in~\tabref{soon}, on the validation unseen split, our model HM3D-FSTTA achieves SR and SPL of 42.44\% and 31.03\%, respectively, while the state-of-the-art method GridMM are 37.46\% and 24.81\%. On the test unseen split, our approach improves the performance of DUET by substantial gains (\emph{e.g.}, $21.42\% \rightarrow 23.23\%$ for SPL). 

\Paragraph{R2R-CE.}
FSTTA also generalizes well on the continuous environment, \emph{i.e.}, R2R-CE dataset, as shown in~\tabref{r2rce}. 
The results indicate that our approach demonstrates superior or comparable performance against other methods. 

\begin{table} [t!]
	\centering
	\caption{Experimental results on SOON dataset. }
	\label{tab:soon}
	\resizebox{\columnwidth}{!}{
		\begin{tabular}{l|*{4}{c}|*{4}{c}}
			\toprule
			\multirow{2}{*}{\textbf{Methods}}
			& \multicolumn{4}{c}{Val Unseen} 
			& \multicolumn{4}{|c}{Test Unseen} \\
			& \cellcolor{blue!25}OSR & \cellcolor{blue!25}SR & \cellcolor{blue!25} SPL & \cellcolor{blue!25} RGSPL 
			& \cellcolor{blue!25}OSR & \cellcolor{blue!25}SR & \cellcolor{blue!25} SPL & \cellcolor{blue!25} RGSPL\\
			\midrule
			GBE~\cite{zhu2021soon}  & 28.54 & 19.52 & 13.34 & 1.16
			& 21.45 & 12.90 & 9.23 & 0.45\\
			GridMM~\cite{wang2023gridmm} & 53.39 & 37.46 & 24.81 & 3.91
			& 48.02 & 36.27 & 21.25 & 4.15\\
			KERM~\cite{li2023kerm}  & 51.62 & 38.05 & 23.16 & 4.04
			& - &- &- &- \\
			AZHP~\cite{gao2023adaptive}  & \textbf{56.19} & 40.71 & 26.58 & \textbf{5.53}  
			& - &- &- &-\\ 
			\midrule
			DUET~\cite{chen2022think} & 50.91 & 36.28 & 22.58 & 3.75
			& 43.00 & 33.44 & 21.42 & 4.17\\
			DUET-FSTTA & \textcolor{red}{52.57} & \textcolor{red}{36.53} & \textcolor{red}{23.82} & \textcolor{red}{3.75}  
			& \textcolor{red}{43.44} & \textcolor{red}{35.34} & \textcolor{red}{23.23} & \textcolor{red}{4.52}\\
			\midrule
			HM3D~\cite{chen2022learning}  & 53.22 & 41.00 & 30.69 & 4.06
			& 47.26 & 40.26 & 28.09 & 5.15\\
			HM3D-FSTTA  & \textcolor{red}{54.19} & \textcolor{red}{\textbf{42.44}} & \textcolor{red}{\textbf{31.03}} & \textcolor{red}{4.93} & \textcolor{red}{\textbf{48.52}} & \textcolor{red}{\textbf{42.02}} & \textcolor{red}{\textbf{28.95}} & \textcolor{red}{\textbf{5.20}} \\
			\bottomrule
		\end{tabular}
	}
\end{table}

\begin{table}[t!]\centering
	\caption{Experimental results on R2R-CE dataset.}
	\label{tab:r2rce}
	\resizebox{0.48\textwidth}{!}{
		\large
		\begin{tabular}{l|*{4}{c}|*{4}{c}}
			\toprule
			\multirow{2}{*}{\textbf{Methods}}
			& \multicolumn{4}{c}{Val Unseen} 
			& \multicolumn{4}{|c}{Test Unseen} \\
			&\cellcolor{red!25}NE $\downarrow$  & \cellcolor{blue!25}OSR & \cellcolor{blue!25}SR & \cellcolor{blue!25} SPL 
			& \cellcolor{red!25}NE $\downarrow$ & \cellcolor{blue!25}OSR & \cellcolor{blue!25}SR & \cellcolor{blue!25} SPL \\
			\midrule
			Seq2Seq~\cite{krantz2020beyond} & 7.37 & 40 & 32 & 30 & 7.91 & 36 & 28 & 25\\
			Sim2Sim~\cite{Krantz2022Sim2SimTF}  & 6.07 & 52 & 43 & 36 & 6.17 & 52 & 44 & 37 \\
			CWP-BERT~\cite{Hong2022BridgingTG}  & 5.74 & 53 & 44 & 39 & 5.89 & 51 & 42 & 36\\    
			DREAMW~\cite{Wang2023DREAMWALKERMP}  & 5.53 & 49 & 59 & 44  & 5.48 & 49 & 57 & 44 \\
			GridMM~\cite{wang2023gridmm}  & 5.11 & 61 & 49 & 41 & 5.64 & 56 & 46 & 39 \\
			ETPNav~\cite{An2023ETPNavET}  & 4.71 & 65 & 57 & 49 & 5.12 & 63 & 55 & 48  \\
			\midrule
			WS-MGMap~\cite{Chen2022WeaklySupervisedMM}  & 6.28 & 48 & 39 & 34 & 7.11 & 45 & 35 & 28  \\
			WS-MGMap-FSTTA & \textcolor{red}{6.16} & \textcolor{red}{49} & \textcolor{red}{40} & \textcolor{red}{35} & \textcolor{blue}{7.62} & \textcolor{red}{46} & \textcolor{red}{37} & \textcolor{red}{28} \\
			\midrule
			DUET~\cite{chen2022think}  & 5.13 & 55 & 46 & 40 & 5.82 & 50 & 42 & 36\\
			DUET-FSTTA  & \textcolor{blue}{5.27} & \textcolor{red}{58} & \textcolor{red}{48} & \textcolor{red}{42} & \textcolor{blue}{5.84} & \textcolor{red}{55} & \textcolor{red}{46} & \textcolor{red}{38}  \\
			\midrule
			BEVBert~\cite{an2023bevbert}  & 4.57 & \textbf{67} & 59 & 50 & \textbf{4.70} & 67 & 59 & 50  \\
			BEVBert-FSTTA &  \textcolor{red}{\textbf{4.39}} &  \textcolor{blue}{65} &  \textcolor{red}{\textbf{60}} & \textcolor{red}{\textbf{51}} & \textcolor{blue}{5.45}  & \textcolor{red}{\textbf{69}}  & \textcolor{red}{\textbf{60}} & \textcolor{red}{\textbf{50}}  \\
			\bottomrule
		\end{tabular}
	}
\end{table}

\begin{figure}[t!]
	\centering
	\includegraphics[width=1\linewidth]{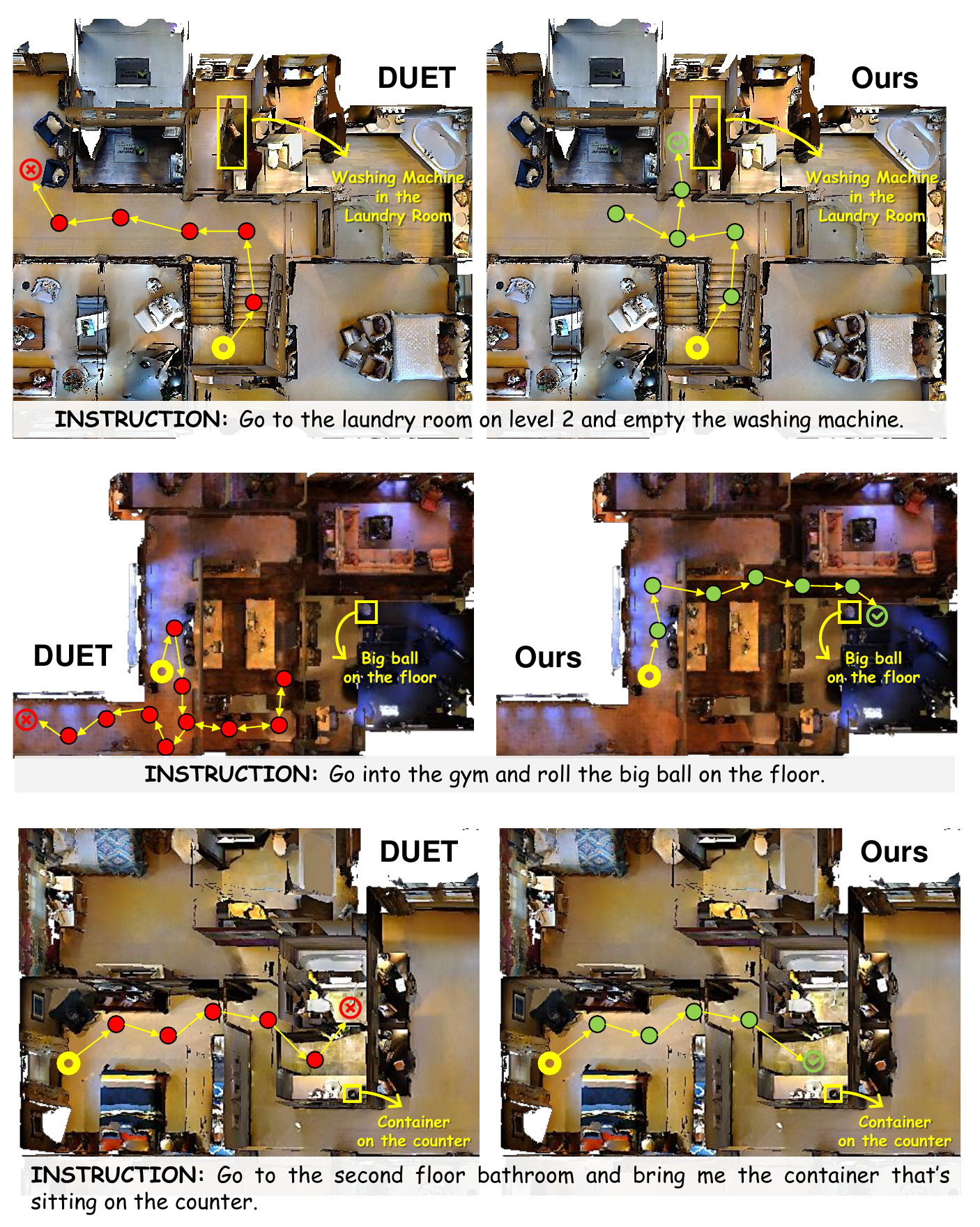}
	\vspace{-6mm}
	\caption{
		Representative visual results on REVERIE validation unseen set. 
		Yellow points denote start locations, while the directed lines with red and green points depict the predicted trajectories with target and incorrect endpoints, respectively.
		With FSTTA, the basic agent (DUET) demonstrates enhanced exploration capabilities, effectively moving towards the correct direction, and succeeds based on the object context and scene layouts.
	}
	\label{fig:qualitative}
	\vspace{-4mm}
\end{figure}

\Paragraph{Qualitative Analysis.}
\figref{qualitative} provides a visualization of the agent's instruction execution process, validating that our proposed FSTTA approach can indeed dynamically enhance the VLN performance of the agent during testing.

\section{Conclusions}
\label{sec:conclus}
This paper explores the feasibility of TTA strategies for online VLN. We propose a fast-slow test-time adaptation method, which performs decomposition-accumulation analysis for both gradients and parameters, achieving a balance between adaptability and stability. 
The encouraging performance is validated in extensive experiments.

\Paragraph{Limitations.} Several limitations 
are noteworthy. 
Firstly, our approach focuses on adapting normalization layers within the model. While these layers are widely employed in deep learning, there are still a few methods that do not utilize them. One viable approach to address this issue is to introduce additional normalization layers to the corresponding models and retrain them using the training data. In the future, we will also explore how our model can update other types of layers.
Secondly, 
In this paper, we simulate the online VLN setting simply by sequentially inputting data from the test set. In the future, we aim to construct a more realistic agent online learning dataset that aligns with practical application scenarios, to better evaluate TTA methods.
Thirdly, compared to the base model, the introduction of TTA inevitably incurs additional computational cost, which is a direction for future improvement. Finally, the frequencies of fast and slow updates are fixed and periodic. Adaptive update invocation strategies is worthy of consideration.



\nocite{langley00}

{
\balance
\bibliography{icml2024bib}

\begin{thebibliography}{96}
\providecommand{\natexlab}[1]{#1}
\providecommand{\url}[1]{\texttt{#1}}
\expandafter\ifx\csname urlstyle\endcsname\relax
  \providecommand{\doi}[1]{doi: #1}\else
  \providecommand{\doi}{doi: \begingroup \urlstyle{rm}\Url}\fi

\bibitem[An et~al.(2021)An, Qi, Huang, Wu, Wang, and Tan]{An2021NeighborviewEM}
An, D., Qi, Y., Huang, Y., Wu, Q., Wang, L., and Tan, T.
\newblock Neighbor-view enhanced model for vision and language navigation.
\newblock In \emph{ACM MM}, pp.\  5101--5109, 2021.

\bibitem[An et~al.(2022)An, Qi, Li, Huang, Wang, Tan, and Shao]{an2023bevbert}
An, D., Qi, Y., Li, Y., Huang, Y., Wang, L., Tan, T., and Shao, J.
\newblock Bevbert: Topo-metric map pre-training for language-guided navigation.
\newblock \emph{arXiv preprint arXiv:2212.04385}, 2022.

\bibitem[An et~al.(2023)An, Wang, Wang, Wang, Huang, He, and
  Wang]{An2023ETPNavET}
An, D., Wang, H., Wang, W., Wang, Z., Huang, Y., He, K., and Wang, L.
\newblock Etpnav: Evolving topological planning for vision-language navigation
  in continuous environments.
\newblock \emph{ArXiv}, abs/2304.03047, 2023.

\bibitem[Anderson et~al.(2018)Anderson, Wu, Teney, Bruce, Johnson,
  S{\"u}nderhauf, Reid, Gould, and Van Den~Hengel]{anderson2018vision}
Anderson, P., Wu, Q., Teney, D., Bruce, J., Johnson, M., S{\"u}nderhauf, N.,
  Reid, I., Gould, S., and Van Den~Hengel, A.
\newblock Vision-and-language navigation: Interpreting visually-grounded
  navigation instructions in real environments.
\newblock In \emph{CVPR}, pp.\  3674--3683, 2018.

\bibitem[Anderson et~al.(2019)Anderson, Shrivastava, Parikh, Batra, and
  Lee]{Anderson2019ChasingGI}
Anderson, P., Shrivastava, A., Parikh, D., Batra, D., and Lee, S.
\newblock Chasing ghosts: Instruction following as bayesian state tracking.
\newblock In \emph{NeurIPS}, volume~32, pp.\  371--381, 2019.

\bibitem[Barzilai \& Borwein(1988)Barzilai and Borwein]{Barzilai1988TwoPointSS}
Barzilai, J. and Borwein, J.~M.
\newblock Two-point step size gradient methods.
\newblock \emph{Ima Journal of Numerical Analysis}, 8:\penalty0 141--148, 1988.

\bibitem[Boudiaf et~al.(2022)Boudiaf, Mueller, Ben~Ayed, and
  Bertinetto]{boudiaf2022parameter}
Boudiaf, M., Mueller, R., Ben~Ayed, I., and Bertinetto, L.
\newblock Parameter-free online test-time adaptation.
\newblock In \emph{CVPR}, pp.\  8344--8353, 2022.

\bibitem[Chen et~al.(2022{\natexlab{a}})Chen, Gao, Meng, Zhang, and
  Liu]{Chen2022ReinforcedSS}
Chen, J., Gao, C., Meng, E., Zhang, Q., and Liu, S.
\newblock Reinforced structured state-evolution for vision-language navigation.
\newblock In \emph{CVPR}, pp.\  15429--15438, 2022{\natexlab{a}}.

\bibitem[Chen et~al.(2020)Chen, Chen, Chuang, V'azquez, and
  Savarese]{Chen2020TopologicalPW}
Chen, K., Chen, J., Chuang, J., V'azquez, M., and Savarese, S.
\newblock Topological planning with transformers for vision-and-language
  navigation.
\newblock In \emph{CVPR}, pp.\  11271--11281, 2020.

\bibitem[Chen et~al.(2022{\natexlab{b}})Chen, Ji, Lin, Zeng, Li, Tan, and
  Gan]{Chen2022WeaklySupervisedMM}
Chen, P., Ji, D., Lin, K.-L.~C., Zeng, R., Li, T.~H., Tan, M., and Gan, C.
\newblock Weakly-supervised multi-granularity map learning for
  vision-and-language navigation.
\newblock In \emph{NeurIPS}, pp.\  38149--38161, 2022{\natexlab{b}}.

\bibitem[Chen et~al.(2021)Chen, Guhur, Schmid, and Laptev]{chen2021history}
Chen, S., Guhur, P.-L., Schmid, C., and Laptev, I.
\newblock History aware multimodal transformer for vision-and-language
  navigation.
\newblock In \emph{NeurIPS}, pp.\  5834--5847, 2021.

\bibitem[Chen et~al.(2022{\natexlab{c}})Chen, Guhur, Tapaswi, Schmid, and
  Laptev]{chen2022learning}
Chen, S., Guhur, P.-L., Tapaswi, M., Schmid, C., and Laptev, I.
\newblock Learning from unlabeled 3d environments for vision-and-language
  navigation.
\newblock In \emph{ECCV}, pp.\  638--655, 2022{\natexlab{c}}.

\bibitem[Chen et~al.(2022{\natexlab{d}})Chen, Guhur, Tapaswi, Schmid, and
  Laptev]{chen2022think}
Chen, S., Guhur, P.-L., Tapaswi, M., Schmid, C., and Laptev, I.
\newblock Think global, act local: Dual-scale graph transformer for
  vision-and-language navigation.
\newblock In \emph{CVPR}, pp.\  16537--16547, 2022{\natexlab{d}}.

\bibitem[Cui et~al.(2023)Cui, Xie, Zhang, Zhang, Yan, and Yin]{cui2023grounded}
Cui, Y., Xie, L., Zhang, Y., Zhang, M., Yan, Y., and Yin, E.
\newblock Grounded entity-landmark adaptive pre-training for
  vision-and-language navigation.
\newblock In \emph{ICCV}, pp.\  12043--12053, 2023.

\bibitem[D{\"o}bler et~al.(2023)D{\"o}bler, Marsden, and
  Yang]{dobler2023robust}
D{\"o}bler, M., Marsden, R.~A., and Yang, B.
\newblock Robust mean teacher for continual and gradual test-time adaptation.
\newblock In \emph{CVPR}, pp.\  7704--7714, 2023.

\bibitem[Dosovitskiy et~al.(2020)Dosovitskiy, Beyer, Kolesnikov, Weissenborn,
  Zhai, Unterthiner, Dehghani, Minderer, Heigold, Gelly,
  et~al.]{dosovitskiy2020image}
Dosovitskiy, A., Beyer, L., Kolesnikov, A., Weissenborn, D., Zhai, X.,
  Unterthiner, T., Dehghani, M., Minderer, M., Heigold, G., Gelly, S., et~al.
\newblock An image is worth 16x16 words: Transformers for image recognition at
  scale.
\newblock In \emph{ICLR}, 2020.

\bibitem[Du et~al.(2018)Du, Czarnecki, Jayakumar, Farajtabar, Pascanu, and
  Lakshminarayanan]{du2018adapting}
Du, Y., Czarnecki, W.~M., Jayakumar, S.~M., Farajtabar, M., Pascanu, R., and
  Lakshminarayanan, B.
\newblock Adapting auxiliary losses using gradient similarity.
\newblock \emph{arXiv preprint arXiv:1812.02224}, 2018.

\bibitem[Fried et~al.(2018)Fried, Hu, Cirik, Rohrbach, Andreas, Morency,
  Berg-Kirkpatrick, Saenko, Klein, and Darrell]{fried2018speaker}
Fried, D., Hu, R., Cirik, V., Rohrbach, A., Andreas, J., Morency, L.-P.,
  Berg-Kirkpatrick, T., Saenko, K., Klein, D., and Darrell, T.
\newblock Speaker-follower models for vision-and-language navigation.
\newblock In \emph{NeurIPS}, volume~31, 2018.

\bibitem[Gao et~al.(2021{\natexlab{a}})Gao, Chen, Liu, Wang, Zhang, and
  Wu]{Gao2021RoomandObjectAK}
Gao, C., Chen, J., Liu, S., Wang, L., Zhang, Q., and Wu, Q.
\newblock Room-and-object aware knowledge reasoning for remote embodied
  referring expression.
\newblock In \emph{CVPR}, pp.\  3063--3072, 2021{\natexlab{a}}.

\bibitem[Gao et~al.(2023{\natexlab{a}})Gao, Peng, Yan, Wang, Yang, Ren, Li, and
  Liu]{gao2023adaptive}
Gao, C., Peng, X., Yan, M., Wang, H., Yang, L., Ren, H., Li, H., and Liu, S.
\newblock Adaptive zone-aware hierarchical planner for vision-language
  navigation.
\newblock In \emph{CVPR}, pp.\  14911--14920, 2023{\natexlab{a}}.

\bibitem[Gao \& Xu(2021)Gao and Xu]{gao2021learning}
Gao, J. and Xu, C.
\newblock Learning video moment retrieval without a single annotated video.
\newblock \emph{IEEE Transactions on Circuits and Systems for Video
  Technology}, 32\penalty0 (3):\penalty0 1646--1657, 2021.

\bibitem[Gao et~al.(2021{\natexlab{b}})Gao, Zhang, and Xu]{gao2020learning}
Gao, J., Zhang, T., and Xu, C.
\newblock Learning to model relationships for zero-shot video classification.
\newblock \emph{IEEE Transactions on Pattern Analysis and Machine
  Intelligence}, 43\penalty0 (10):\penalty0 3476--3491, 2021{\natexlab{b}}.

\bibitem[Gao et~al.(2023{\natexlab{b}})Gao, Chen, and Xu]{gao2023vectorized}
Gao, J., Chen, M., and Xu, C.
\newblock Vectorized evidential learning for weakly-supervised temporal action
  localization.
\newblock \emph{IEEE transactions on pattern analysis and machine
  intelligence}, 45:\penalty0 15949 -- 15963, 2023{\natexlab{b}}.

\bibitem[Georgakis et~al.(2022)Georgakis, Schmeckpeper, Wanchoo, Dan,
  Miltsakaki, Roth, and Daniilidis]{Georgakis2022CrossmodalML}
Georgakis, G., Schmeckpeper, K., Wanchoo, K., Dan, S., Miltsakaki, E., Roth,
  D., and Daniilidis, K.
\newblock Cross-modal map learning for vision and language navigation.
\newblock In \emph{CVPR}, pp.\  15439--15449, 2022.

\bibitem[Gong et~al.(2022)Gong, Jeong, Kim, Kim, Shin, and Lee]{gong2022note}
Gong, T., Jeong, J., Kim, T., Kim, Y., Shin, J., and Lee, S.-J.
\newblock Note: Robust continual test-time adaptation against temporal
  correlation.
\newblock In \emph{NeurIPS}, pp.\  27253--27266, 2022.

\bibitem[Gu et~al.(2022)Gu, Stefani, Wu, Thomason, and Wang]{gu2022vision}
Gu, J., Stefani, E., Wu, Q., Thomason, J., and Wang, X.
\newblock Vision-and-language navigation: A survey of tasks, methods, and
  future directions.
\newblock In \emph{ACL}, pp.\  7606--7623, 2022.

\bibitem[Guhur et~al.(2021)Guhur, Tapaswi, Chen, Laptev, and
  Schmid]{Guhur2021AirbertIP}
Guhur, P.-L., Tapaswi, M., Chen, S., Laptev, I., and Schmid, C.
\newblock Airbert: In-domain pretraining for vision-and-language navigation.
\newblock In \emph{ICCV}, pp.\  1614--1623, 2021.

\bibitem[Hao et~al.(2020)Hao, Li, Li, Carin, and Gao]{hao2020towards}
Hao, W., Li, C., Li, X., Carin, L., and Gao, J.
\newblock Towards learning a generic agent for vision-and-language navigation
  via pre-training.
\newblock In \emph{CVPR}, pp.\  13137--13146, 2020.

\bibitem[Hong et~al.(2020)Hong, Rodriguez-Opazo, Qi, Wu, and
  Gould]{Hong2020LanguageAV}
Hong, Y., Rodriguez-Opazo, C., Qi, Y., Wu, Q., and Gould, S.
\newblock Language and visual entity relationship graph for agent navigation.
\newblock In \emph{NeurIPS}, pp.\  7685--7696, 2020.

\bibitem[Hong et~al.(2021)Hong, Wu, Qi, Rodriguez-Opazo, and
  Gould]{hong2021vln}
Hong, Y., Wu, Q., Qi, Y., Rodriguez-Opazo, C., and Gould, S.
\newblock Vln bert: A recurrent vision-and-language bert for navigation.
\newblock In \emph{CVPR}, pp.\  1643--1653, 2021.

\bibitem[Hong et~al.(2022)Hong, Wang, Wu, and Gould]{Hong2022BridgingTG}
Hong, Y., Wang, Z., Wu, Q., and Gould, S.
\newblock Bridging the gap between learning in discrete and continuous
  environments for vision-and-language navigation.
\newblock In \emph{CVPR}, pp.\  15418--15428, 2022.

\bibitem[Hu et~al.(2024)Hu, Gao, Dong, Fan, and Liu]{hu2023exploring}
Hu, Y., Gao, J., Dong, J., Fan, B., and Liu, H.
\newblock Exploring rich semantics for open-set action recognition.
\newblock \emph{IEEE Transactions on Multimedia}, 26:\penalty0 5410 -- 5421,
  2024.

\bibitem[Huo et~al.(2023)Huo, Sun, Jiang, Lin, and Fu]{huo2023geovln}
Huo, J., Sun, Q., Jiang, B., Lin, H., and Fu, Y.
\newblock Geovln: Learning geometry-enhanced visual representation with slot
  attention for vision-and-language navigation.
\newblock In \emph{CVPR}, pp.\  23212--23221, 2023.

\bibitem[Irshad et~al.(2021)Irshad, Mithun, Seymour, Chiu, Samarasekera, and
  Kumar]{Irshad2021SemanticallyawareSR}
Irshad, M.~Z., Mithun, N.~C., Seymour, Z., Chiu, H.-P., Samarasekera, S., and
  Kumar, R.
\newblock Semantically-aware spatio-temporal reasoning agent for
  vision-and-language navigation in continuous environments.
\newblock In \emph{ICPR}, pp.\  4065--4071, 2021.

\bibitem[Kamath et~al.(2023)Kamath, Anderson, Wang, Koh, Ku, Waters, Yang,
  Baldridge, and Parekh]{kamath2023new}
Kamath, A., Anderson, P., Wang, S., Koh, J.~Y., Ku, A., Waters, A., Yang, Y.,
  Baldridge, J., and Parekh, Z.
\newblock A new path: Scaling vision-and-language navigation with synthetic
  instructions and imitation learning.
\newblock In \emph{CVPR}, pp.\  10813--10823, 2023.

\bibitem[Ke et~al.(2019)Ke, Li, Bisk, Holtzman, Gan, Liu, Gao, Choi, and
  Srinivasa]{ke2019tactical}
Ke, L., Li, X., Bisk, Y., Holtzman, A., Gan, Z., Liu, J., Gao, J., Choi, Y.,
  and Srinivasa, S.
\newblock Tactical rewind: Self-correction via backtracking in
  vision-and-language navigation.
\newblock In \emph{CVPR}, pp.\  6741--6749, 2019.

\bibitem[Krantz \& Lee(2022)Krantz and Lee]{Krantz2022Sim2SimTF}
Krantz, J. and Lee, S.
\newblock Sim-2-sim transfer for vision-and-language navigation in continuous
  environments.
\newblock In \emph{ECCV}, pp.\  588--603, 2022.

\bibitem[Krantz et~al.(2020)Krantz, Wijmans, Majumdar, Batra, and
  Lee]{krantz2020beyond}
Krantz, J., Wijmans, E., Majumdar, A., Batra, D., and Lee, S.
\newblock Beyond the nav-graph: Vision-and-language navigation in continuous
  environments.
\newblock In \emph{ECCV}, pp.\  104--120, 2020.

\bibitem[Krantz et~al.(2021)Krantz, Gokaslan, Batra, Lee, and
  Maksymets]{Krantz2021WaypointMF}
Krantz, J., Gokaslan, A., Batra, D., Lee, S., and Maksymets, O.
\newblock Waypoint models for instruction-guided navigation in continuous
  environments.
\newblock In \emph{ICCV}, pp.\  15142--15151, 2021.

\bibitem[Ku et~al.(2020)Ku, Anderson, Patel, Ie, and Baldridge]{ku2020room}
Ku, A., Anderson, P., Patel, R., Ie, E., and Baldridge, J.
\newblock Room-across-room: Multilingual vision-and-language navigation with
  dense spatiotemporal grounding.
\newblock In \emph{EMNLP}, pp.\  4392--4412, 2020.

\bibitem[Lee et~al.(2023)Lee, Das, Choo, and Choi]{lee2023towards}
Lee, J., Das, D., Choo, J., and Choi, S.
\newblock Towards open-set test-time adaptation utilizing the wisdom of crowds
  in entropy minimization.
\newblock In \emph{ICCV}, pp.\  16380--16389, 2023.

\bibitem[Lew et~al.(2023)Lew, Son, and Chang]{lew2023gradient}
Lew, B., Son, D., and Chang, B.
\newblock Gradient estimation for unseen domain risk minimization with
  pre-trained models.
\newblock In \emph{ICCV}, pp.\  4436--4446, 2023.

\bibitem[Li \& Bansal(2023)Li and Bansal]{li2023improving}
Li, J. and Bansal, M.
\newblock Improving vision-and-language navigation by generating future-view
  image semantics.
\newblock In \emph{CVPR}, pp.\  10803--10812, 2023.

\bibitem[Li et~al.(2022)Li, Tan, and Bansal]{li2022envedit}
Li, J., Tan, H., and Bansal, M.
\newblock Envedit: Environment editing for vision-and-language navigation.
\newblock In \emph{CVPR}, pp.\  6741--6749, 2022.

\bibitem[Li et~al.(2019)Li, Li, Xia, Bisk, Celikyilmaz, Gao, Smith, and
  Choi]{Li2019RobustNW}
Li, X., Li, C., Xia, Q., Bisk, Y., Celikyilmaz, A., Gao, J., Smith, N.~A., and
  Choi, Y.
\newblock Robust navigation with language pretraining and stochastic sampling.
\newblock In \emph{EMNLP}, pp.\  1494--1499, 2019.

\bibitem[Li et~al.(2023)Li, Wang, Yang, Wang, and Jiang]{li2023kerm}
Li, X., Wang, Z., Yang, J., Wang, Y., and Jiang, S.
\newblock Kerm: Knowledge enhanced reasoning for vision-and-language
  navigation.
\newblock In \emph{CVPR}, pp.\  2583--2592, 2023.

\bibitem[Liang et~al.(2023)Liang, He, and Tan]{liang2023comprehensive}
Liang, J., He, R., and Tan, T.
\newblock A comprehensive survey on test-time adaptation under distribution
  shifts.
\newblock \emph{arXiv preprint arXiv:2303.15361}, 2023.

\bibitem[Lim et~al.(2023)Lim, Kim, Choo, and Choi]{lim2022ttn}
Lim, H., Kim, B., Choo, J., and Choi, S.
\newblock Ttn: A domain-shift aware batch normalization in test-time
  adaptation.
\newblock In \emph{ICLR}, 2023.

\bibitem[Lin et~al.(2022)Lin, Jiang, Cai, Qu, Haffari, and
  Yuan]{lin2022multimodal}
Lin, C., Jiang, Y., Cai, J., Qu, L., Haffari, G., and Yuan, Z.
\newblock Multimodal transformer with variable-length memory for
  vision-and-language navigation.
\newblock In \emph{ECCV}, pp.\  380--397, 2022.

\bibitem[Lin et~al.(2023)Lin, Mirza, Kozinski, Possegger, Kuehne, and
  Bischof]{lin2023video}
Lin, W., Mirza, M.~J., Kozinski, M., Possegger, H., Kuehne, H., and Bischof, H.
\newblock Video test-time adaptation for action recognition.
\newblock In \emph{CVPR}, pp.\  22952--22961, 2023.

\bibitem[Lin et~al.(2021)Lin, Li, and Yu]{Lin2021SceneIntuitiveAF}
Lin, X., Li, G., and Yu, Y.
\newblock Scene-intuitive agent for remote embodied visual grounding.
\newblock In \emph{CVPR}, pp.\  7032--7041, 2021.

\bibitem[Liu et~al.(2021)Liu, Zhu, Chang, Liang, and
  Shen]{Liu2021VisionLanguageNW}
Liu, C., Zhu, F., Chang, X., Liang, X., and Shen, Y.-D.
\newblock Vision-language navigation with random environmental mixup.
\newblock In \emph{ICCV}, pp.\  1624--1634, 2021.

\bibitem[Liu et~al.(2024)Liu, Yang, Jia, Lu, Guo, Xue, and
  Zhang]{Liu2023ViDAHV}
Liu, J., Yang, S., Jia, P., Lu, M., Guo, Y., Xue, W., and Zhang, S.
\newblock Vida: Homeostatic visual domain adapter for continual test time
  adaptation.
\newblock In \emph{ICLR}, 2024.

\bibitem[Liu et~al.(2023)Liu, Wang, Wang, and Yang]{Liu2023BirdsEyeViewSG}
Liu, R., Wang, X., Wang, W., and Yang, Y.
\newblock Bird's-eye-view scene graph for vision-language navigation.
\newblock In \emph{ICCV}, pp.\  10968--10980, 2023.

\bibitem[Loshchilov \& Hutter(2017)Loshchilov and
  Hutter]{Loshchilov2017DecoupledWD}
Loshchilov, I. and Hutter, F.
\newblock Decoupled weight decay regularization.
\newblock In \emph{ICLR}, 2017.

\bibitem[Lu et~al.(2022)Lu, Zhang, Nie, Feng, Xu, Wang, and
  Wang]{lu2022anticipating}
Lu, Y., Zhang, H., Nie, P., Feng, W., Xu, W., Wang, X.~E., and Wang, W.~Y.
\newblock Anticipating the unseen discrepancy for vision and language
  navigation.
\newblock \emph{arXiv preprint arXiv:2209.04725}, 2022.

\bibitem[Ma et~al.(2019{\natexlab{a}})Ma, Lu, Wu, AlRegib, Kira, Socher, and
  Xiong]{ma2019self}
Ma, C.-Y., Lu, J., Wu, Z., AlRegib, G., Kira, Z., Socher, R., and Xiong, C.
\newblock Self-monitoring navigation agent via auxiliary progress estimation.
\newblock In \emph{ICLR}, 2019{\natexlab{a}}.

\bibitem[Ma et~al.(2019{\natexlab{b}})Ma, Wu, AlRegib, Xiong, and
  Kira]{ma2019regretful}
Ma, C.-Y., Wu, Z., AlRegib, G., Xiong, C., and Kira, Z.
\newblock The regretful agent: Heuristic-aided navigation through progress
  estimation.
\newblock In \emph{CVPR}, pp.\  6732--6740, 2019{\natexlab{b}}.

\bibitem[Mansilla et~al.(2021)Mansilla, Echeveste, Milone, and
  Ferrante]{mansilla2021domain}
Mansilla, L., Echeveste, R., Milone, D.~H., and Ferrante, E.
\newblock Domain generalization via gradient surgery.
\newblock In \emph{ICCV}, pp.\  6630--6638, 2021.

\bibitem[Mirza et~al.(2022)Mirza, Micorek, Possegger, and
  Bischof]{mirza2022norm}
Mirza, M.~J., Micorek, J., Possegger, H., and Bischof, H.
\newblock The norm must go on: Dynamic unsupervised domain adaptation by
  normalization.
\newblock In \emph{CVPR}, pp.\  14765--14775, 2022.

\bibitem[Nguyen et~al.(2019)Nguyen, Dey, Brockett, and Dolan]{nguyen2019vision}
Nguyen, K., Dey, D., Brockett, C., and Dolan, B.
\newblock Vision-based navigation with language-based assistance via imitation
  learning with indirect intervention.
\newblock In \emph{CVPR}, pp.\  12527--12537, 2019.

\bibitem[Niu et~al.(2022)Niu, Wu, Zhang, Chen, Zheng, Zhao, and
  Tan]{niu2022efficient}
Niu, S., Wu, J., Zhang, Y., Chen, Y., Zheng, S., Zhao, P., and Tan, M.
\newblock Efficient test-time model adaptation without forgetting.
\newblock In \emph{ICML}, pp.\  16888--16905, 2022.

\bibitem[Niu et~al.(2023)Niu, Wu, Zhang, Wen, Chen, Zhao, and
  Tan]{niu2023towards}
Niu, S., Wu, J., Zhang, Y., Wen, Z., Chen, Y., Zhao, P., and Tan, M.
\newblock Towards stable test-time adaptation in dynamic wild world.
\newblock In \emph{ICLR}, 2023.

\bibitem[Qi et~al.(2020)Qi, Wu, Anderson, Wang, Wang, Shen, and
  Hengel]{qi2020reverie}
Qi, Y., Wu, Q., Anderson, P., Wang, X., Wang, W.~Y., Shen, C., and Hengel, A.
  v.~d.
\newblock Reverie: Remote embodied visual referring expression in real indoor
  environments.
\newblock In \emph{CVPR}, pp.\  9982--9991, 2020.

\bibitem[Qi et~al.(2021)Qi, Pan, Hong, Yang, van~den Hengel, and
  Wu]{Qi2021TheRT}
Qi, Y., Pan, Z., Hong, Y., Yang, M.-H., van~den Hengel, A., and Wu, Q.
\newblock The road to know-where: An and-room informed sequential bert for
  indoor vision-language navigation.
\newblock In \emph{ICCV}, pp.\  1635--1644, 2021.

\bibitem[Qiao et~al.(2022)Qiao, Qi, Hong, Yu, Wang, and Wu]{Qiao2022HOPHA}
Qiao, Y., Qi, Y., Hong, Y., Yu, Z., Wang, P., and Wu, Q.
\newblock Hop: History-and-order aware pretraining for vision-and-language
  navigation.
\newblock In \emph{CVPR}, pp.\  15397--15406, 2022.

\bibitem[Qiao et~al.(2023{\natexlab{a}})Qiao, Qi, Yu, Liu, and
  Wu]{Qiao2023MarchIC}
Qiao, Y., Qi, Y., Yu, Z., Liu, J., and Wu, Q.
\newblock March in chat: Interactive prompting for remote embodied referring
  expression.
\newblock In \emph{ICCV}, pp.\  15758--15767, 2023{\natexlab{a}}.

\bibitem[Qiao et~al.(2023{\natexlab{b}})Qiao, Yu, and Wu]{qiao2023vlnpetl}
Qiao, Y., Yu, Z., and Wu, Q.
\newblock Vln-petl: Parameter-efficient transfer learning for
  vision-and-language navigation.
\newblock In \emph{ICCV}, pp.\  15443--15452, 2023{\natexlab{b}}.

\bibitem[Rame et~al.(2022)Rame, Dancette, and Cord]{rame2022fishr}
Rame, A., Dancette, C., and Cord, M.
\newblock Fishr: Invariant gradient variances for out-of-distribution
  generalization.
\newblock In \emph{ICML}, pp.\  18347--18377, 2022.

\bibitem[Ramrakhya et~al.(2022)Ramrakhya, Undersander, Batra, and
  Das]{ramrakhya2022habitat}
Ramrakhya, R., Undersander, E., Batra, D., and Das, A.
\newblock Habitat-web: Learning embodied object-search strategies from human
  demonstrations at scale.
\newblock In \emph{CVPR}, pp.\  5173--5183, 2022.

\bibitem[Raychaudhuri et~al.(2021)Raychaudhuri, Wani, Patel, Jain, and
  Chang]{Raychaudhuri2021LanguageAlignedW}
Raychaudhuri, S., Wani, S., Patel, S., Jain, U., and Chang, A.~X.
\newblock Language-aligned waypoint (law) supervision for vision-and-language
  navigation in continuous environments.
\newblock In \emph{EMNLP}, pp.\  4018--4028, 2021.

\bibitem[Shi et~al.(2021)Shi, Seely, Torr, Siddharth, Hannun, Usunier, and
  Synnaeve]{shi2021gradient}
Shi, Y., Seely, J., Torr, P., Siddharth, N., Hannun, A., Usunier, N., and
  Synnaeve, G.
\newblock Gradient matching for domain generalization.
\newblock In \emph{ICLR}, 2021.

\bibitem[Shlens(2014)]{shlens2014tutorial}
Shlens, J.
\newblock A tutorial on principal component analysis.
\newblock \emph{arXiv preprint arXiv:1404.1100}, 2014.

\bibitem[Song et~al.(2023)Song, Lee, Kweon, and Choi]{song2023ecotta}
Song, J., Lee, J., Kweon, I.~S., and Choi, S.
\newblock Ecotta: Memory-efficient continual test-time adaptation via
  self-distilled regularization.
\newblock In \emph{CVPR}, pp.\  11920--11929, 2023.

\bibitem[Sun et~al.(2020)Sun, Wang, Liu, Miller, Efros, and Hardt]{sun2020test}
Sun, Y., Wang, X., Liu, Z., Miller, J., Efros, A., and Hardt, M.
\newblock Test-time training with self-supervision for generalization under
  distribution shifts.
\newblock In \emph{ICML}, pp.\  9229--9248, 2020.

\bibitem[Tan et~al.(2019)Tan, Yu, and Bansal]{tan2019learning}
Tan, H., Yu, L., and Bansal, M.
\newblock Learning to navigate unseen environments: Back translation with
  environmental dropout.
\newblock In \emph{Proceedings of NAACL-HLT}, pp.\  2610--2621, 2019.

\bibitem[Tang et~al.(2023)Tang, Zhang, Xu, Chen, Cheng, Leng, Guo, and
  He]{tang2023neuro}
Tang, Y., Zhang, C., Xu, H., Chen, S., Cheng, J., Leng, L., Guo, Q., and He, Z.
\newblock Neuro-modulated hebbian learning for fully test-time adaptation.
\newblock In \emph{CVPR}, pp.\  3728--3738, 2023.

\bibitem[Tian et~al.(2023)Tian, He, Dai, Ma, Liu, and Kira]{tian2023trainable}
Tian, J., He, Z., Dai, X., Ma, C.-Y., Liu, Y.-C., and Kira, Z.
\newblock Trainable projected gradient method for robust fine-tuning.
\newblock In \emph{CVPR}, pp.\  7836--7845, 2023.

\bibitem[Wang et~al.(2021)Wang, Shelhamer, Liu, Olshausen, and
  Darrell]{wang2020tent}
Wang, D., Shelhamer, E., Liu, S., Olshausen, B., and Darrell, T.
\newblock Tent: Fully test-time adaptation by entropy minimization.
\newblock In \emph{ICLR}, 2021.

\bibitem[Wang et~al.(2023{\natexlab{a}})Wang, Liang, Gool, and
  Wang]{Wang2023DREAMWALKERMP}
Wang, H., Liang, W., Gool, L.~V., and Wang, W.
\newblock Dreamwalker: Mental planning for continuous vision-language
  navigation.
\newblock In \emph{ICCV}, pp.\  10873--10883, 2023{\natexlab{a}}.

\bibitem[Wang et~al.(2023{\natexlab{b}})Wang, Zhang, Lei, and
  Zhang]{wang2023sharpness}
Wang, P., Zhang, Z., Lei, Z., and Zhang, L.
\newblock Sharpness-aware gradient matching for domain generalization.
\newblock In \emph{CVPR}, pp.\  3769--3778, 2023{\natexlab{b}}.

\bibitem[Wang et~al.(2022{\natexlab{a}})Wang, Fink, Van~Gool, and
  Dai]{wang2022continual}
Wang, Q., Fink, O., Van~Gool, L., and Dai, D.
\newblock Continual test-time domain adaptation.
\newblock In \emph{CVPR}, pp.\  7201--7211, 2022{\natexlab{a}}.

\bibitem[Wang et~al.(2022{\natexlab{b}})Wang, Montgomery, Orbay, Birodkar,
  Faust, Gur, Jaques, Waters, Baldridge, and Anderson]{wang2022less}
Wang, S., Montgomery, C., Orbay, J., Birodkar, V., Faust, A., Gur, I., Jaques,
  N., Waters, A., Baldridge, J., and Anderson, P.
\newblock Less is more: Generating grounded navigation instructions from
  landmarks.
\newblock In \emph{CVPR}, pp.\  15428--15438, 2022{\natexlab{b}}.

\bibitem[Wang et~al.(2023{\natexlab{c}})Wang, Wu, Yao, and
  Wang]{Wang2023GraphBE}
Wang, T., Wu, Z., Yao, F., and Wang, D.
\newblock Graph based environment representation for vision-and-language
  navigation in continuous environments.
\newblock \emph{ArXiv}, abs/2301.04352, 2023{\natexlab{c}}.

\bibitem[Wang et~al.(2019)Wang, Huang, Celikyilmaz, Gao, Shen, Wang, Wang, and
  Zhang]{wang2019reinforced}
Wang, X., Huang, Q., Celikyilmaz, A., Gao, J., Shen, D., Wang, Y.-F., Wang,
  W.~Y., and Zhang, L.
\newblock Reinforced cross-modal matching and self-supervised imitation
  learning for vision-language navigation.
\newblock In \emph{CVPR}, pp.\  6629--6638, 2019.

\bibitem[Wang et~al.(2023{\natexlab{d}})Wang, Wang, Shao, and
  Yang]{Wang2023LANAAL}
Wang, X., Wang, W., Shao, J., and Yang, Y.
\newblock Lana: A language-capable navigator for instruction following and
  generation.
\newblock In \emph{CVPR}, pp.\  19048--19058, 2023{\natexlab{d}}.

\bibitem[Wang et~al.(2023{\natexlab{e}})Wang, Grigsby, and Qi]{wang2022pgrad}
Wang, Z., Grigsby, J., and Qi, Y.
\newblock Pgrad: Learning principal gradients for domain generalization.
\newblock In \emph{ICLR}, 2023{\natexlab{e}}.

\bibitem[Wang et~al.(2023{\natexlab{f}})Wang, Li, Yang, Liu, and
  Jiang]{wang2023gridmm}
Wang, Z., Li, X., Yang, J., Liu, Y., and Jiang, S.
\newblock Gridmm: Grid memory map for vision-and-language navigation.
\newblock In \emph{ICCV}, pp.\  15625--15636, 2023{\natexlab{f}}.

\bibitem[Yang et~al.(2023)Yang, Majumdar, and Lee]{yang2023behavioral}
Yang, Z., Majumdar, A., and Lee, S.
\newblock Behavioral analysis of vision-and-language navigation agents.
\newblock In \emph{CVPR}, pp.\  2574--2582, 2023.

\bibitem[Yi et~al.(2023)Yi, YANG, Wang, Li, Tan, and Kot]{yi2022temporal}
Yi, C., YANG, S., Wang, Y., Li, H., Tan, Y.-p., and Kot, A.
\newblock Temporal coherent test time optimization for robust video
  classification.
\newblock In \emph{ICLR}, 2023.

\bibitem[Yu et~al.(2020)Yu, Kumar, Gupta, Levine, Hausman, and
  Finn]{yu2020gradient}
Yu, T., Kumar, S., Gupta, A., Levine, S., Hausman, K., and Finn, C.
\newblock Gradient surgery for multi-task learning.
\newblock In \emph{NeurIPS}, volume~33, pp.\  5824--5836, 2020.

\bibitem[Yuan et~al.(2023)Yuan, Xie, and Li]{yuan2023robust}
Yuan, L., Xie, B., and Li, S.
\newblock Robust test-time adaptation in dynamic scenarios.
\newblock In \emph{CVPR}, pp.\  15922--15932, 2023.

\bibitem[Zhang et~al.(2022)Zhang, Levine, and Finn]{zhang2022memo}
Zhang, M., Levine, S., and Finn, C.
\newblock Memo: Test time robustness via adaptation and augmentation.
\newblock In \emph{NeurIPS}, pp.\  38629--38642, 2022.

\bibitem[Zhao et~al.(2023)Zhao, Chen, and Xia]{zhao2023delta}
Zhao, B., Chen, C., and Xia, S.-T.
\newblock Delta: Degradation-free fully test-time adaptation.
\newblock In \emph{ICLR}, 2023.

\bibitem[Zhu et~al.(2020)Zhu, Zhu, Chang, and Liang]{zhu2020vision}
Zhu, F., Zhu, Y., Chang, X., and Liang, X.
\newblock Vision-language navigation with self-supervised auxiliary reasoning
  tasks.
\newblock In \emph{CVPR}, pp.\  10012--10022, 2020.

\bibitem[Zhu et~al.(2021)Zhu, Liang, Zhu, Yu, Chang, and Liang]{zhu2021soon}
Zhu, F., Liang, X., Zhu, Y., Yu, Q., Chang, X., and Liang, X.
\newblock Soon: Scenario oriented object navigation with graph-based
  exploration.
\newblock In \emph{CVPR}, pp.\  12689--12699, 2021.

\end{thebibliography}
\bibliographystyle{icml2024}
}



\end{document}